\documentclass{aifrontiers}

\shorttitle{WebGym: Scaling Training Environments for Visual Web Agents}

\renewcommand{\weblink}{https://webgym-website.github.io/webgym.github.io/}
\renewcommand{\hflink}{https://huggingface.co/datasets/microsoft/webgym_tasks/}
\newcommand{\hfdisplay}{microsoft/webgym\_tasks}
\renewcommand{\ghlink}{https://github.com/microsoft/webgym}
\newcommand{\ghdisplay}{microsoft/webgym}

\usepackage[utf8]{inputenc}
\usepackage{listings}
\tcbset{colback=white,colframe=black!20,boxrule=0.4pt,arc=2mm}

\lstdefinestyle{promptroman}{
  basicstyle=\footnotesize\rmfamily,
  columns=fullflexible,   % <<< fixes the "stretched letters"
  keepspaces=true,
  breaklines=true,
  breakatwhitespace=true
}

\usepackage{booktabs} % for \toprule etc.
\usepackage{capt-of}  % for \captionof inside minipage
\usepackage{algorithm}
\usepackage{algorithmicx}
\usepackage{algcompatible}
\usepackage{algpseudocode}
\usepackage{bm}
\usepackage{hyperref}
\usepackage{cuted}
\usepackage{array} % for m{..} column types
\usepackage{subcaption}

\usepackage{adjustbox}
\usepackage{multirow}
\usepackage{longtable}  % For multi-page tables
\usepackage{wrapfig}
\usepackage{pifont}

\usepackage{tcolorbox}
\usepackage{listings}

\newcommand{\xmark}{\ding{55}}
\newcommand{\cmark}{\ding{51}}   % ✓
\usepackage{pgf}
\usepackage{colortbl}
\usepackage{highlight}  % Load your custom package
\usepackage{tikz}
\usepackage{multirow}

\AtBeginDocument{
  \setlength{\droptitle}{-20pt}
}
\tcbset{
  before skip=6pt,
  after skip=8pt
}

\titlespacing*{\section}{0pt}{8pt plus 2pt minus 2pt}{4pt plus 2pt minus 2pt}
\titlespacing*{\subsection}{0pt}{6pt plus 2pt minus 2pt}{3pt plus 2pt minus 2pt}

\newcommand{\uiuclogo}{\raisebox{-0.2ex}{\includegraphics[height=1.2em]{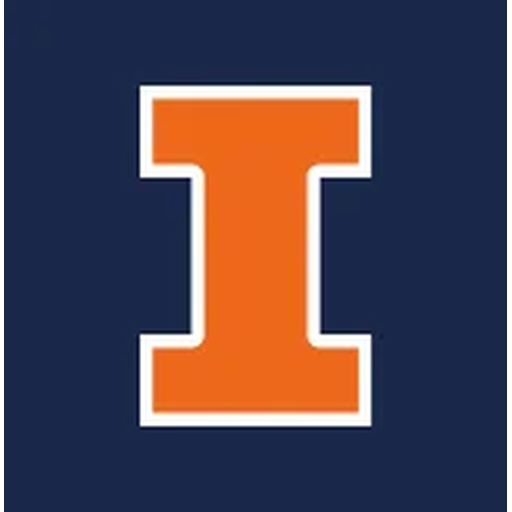}}}
\newcommand{\cmulogo}{\raisebox{-0.2ex}{\includegraphics[height=1.2em]{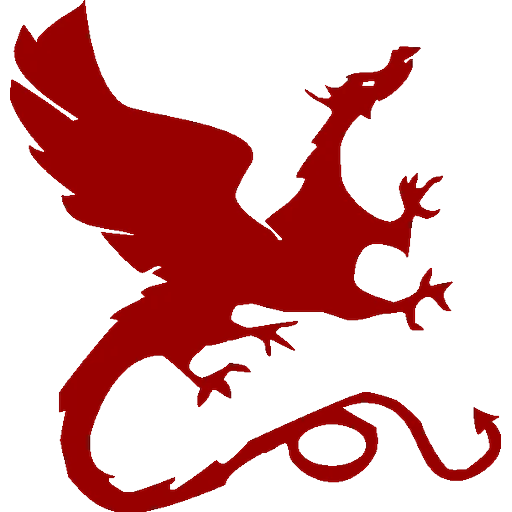}}}

\usepackage[capitalize,noabbrev]{cleveref}

\newcommand{\mslogo}{\includegraphics[height=2.1em]{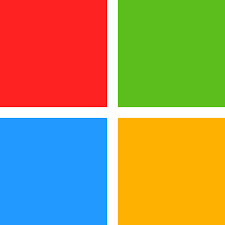}}
\newcommand{\mslogosmall}{%
  \raisebox{-0.2ex}{\includegraphics[height=1.2em]{logo/ms-logo.png}}%
}

\definecolor{successgreen}{RGB}{34, 120, 64}
\definecolor{aifrontiersgray}{HTML}{4D4D4D}
\definecolor{shorttitlecolor}{HTML}{000000}
\definecolor{mygreen}{RGB}{0, 128, 0}       % Classic "green"
\definecolor{forestgreen}{RGB}{34, 139, 34} % Forest green
\definecolor{darkgreen}{RGB}{0, 100, 0}     % Dark green

\makeatletter
\fancypagestyle{firstpage}{%
  \fancyhf{}%
  \fancyhead[L]{}%
  \fancyhead[C]{%
    \raisebox{0pt}[0pt][0pt]{\mslogo}%
    \hspace{0.7em}%
    \raisebox{0.35em}{\sffamily\bfseries\Large\textcolor{aifrontiersgray}{AI Frontiers}}%
  }%
  \fancyhead[R]{\sffamily\small\@date}%
  \renewcommand{\headrulewidth}{1pt}%
  \renewcommand{\headrule}{%
    {\color[HTML]{4D4D4D}\hrule\@height\headrulewidth\@width\headwidth\vskip-\headrulewidth}%
  }%
}
\makeatother

\usepackage{amsmath}
\usepackage{amssymb}
\usepackage{mathtools}
\usepackage{dsfont}  % Changed from bbm to dsfont
\DeclareMathOperator*{\argmax}{arg\,max}

\renewcommand{\AIFRenderLinks}{%
  \vspace{-0.5cm}%
  \begin{center}
    \small
    \href{\weblink}{\weblink}\\[0.3em]%
    \huggingfaceicon\ \href{\hflink}{\hfdisplay}%
    \hspace{1.5em}%
    \githubicon\ \href{\ghlink}{\ghdisplay}%
  \end{center}
  \vspace{-0.4cm}
}

\newcommand{\webgymlogo}{%
  \raisebox{-0.3\baselineskip}{\includegraphics[height=1.3\baselineskip]{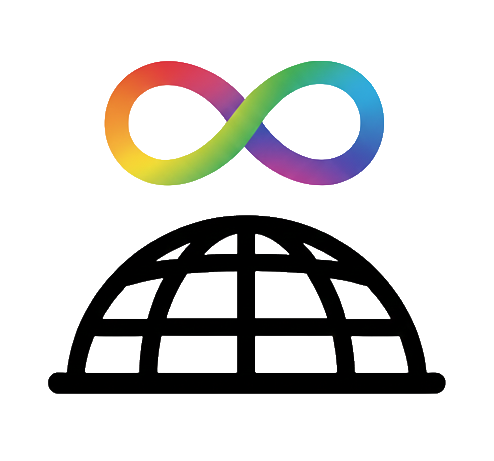}}%
}
\title{%
  \texorpdfstring{\webgymlogo\hspace{0.15em}}{}%
  WebGym: Scaling Training Environments for \\Visual Web Agents with Realistic Tasks
}

\author{
  \textbf{Hao Bai}$^{1,2}$\textsuperscript{\textcolor{shorttitlecolor}{\textdagger}} \quad \textbf{Alexey Taymanov}$^{1}$ \\
  \textbf{Tong Zhang}$^{2}$ \quad \textbf{Aviral Kumar}$^{3}$ \quad \textbf{Spencer Whitehead}$^{1}$ \\[0.5em] 
  $^{1}$ \mslogosmall{} Microsoft \quad $^{2}$ \uiuclogo{} UIUC \quad $^{3}$ \cmulogo{} CMU \\
  \textsuperscript{\textcolor{shorttitlecolor}{\textdagger}}\textcolor{shorttitlecolor}{Work partly done during internship at Microsoft.}
}

\date{\today}

\newcommand{\ours}[1]{\textit{WebGym}}

\begin{document}

% Render title and authors first (manually to control order)
% \makeatletter
\thispagestyle{firstpage}
\begin{center}
% {\LARGE\bfseries\textcolor{shorttitlecolor}{\@title}\par}

\maketitle
\end{center}
\makeatother

% Teaser figure
\vspace{-1em}
\begin{center}
\includegraphics[width=0.98\textwidth]{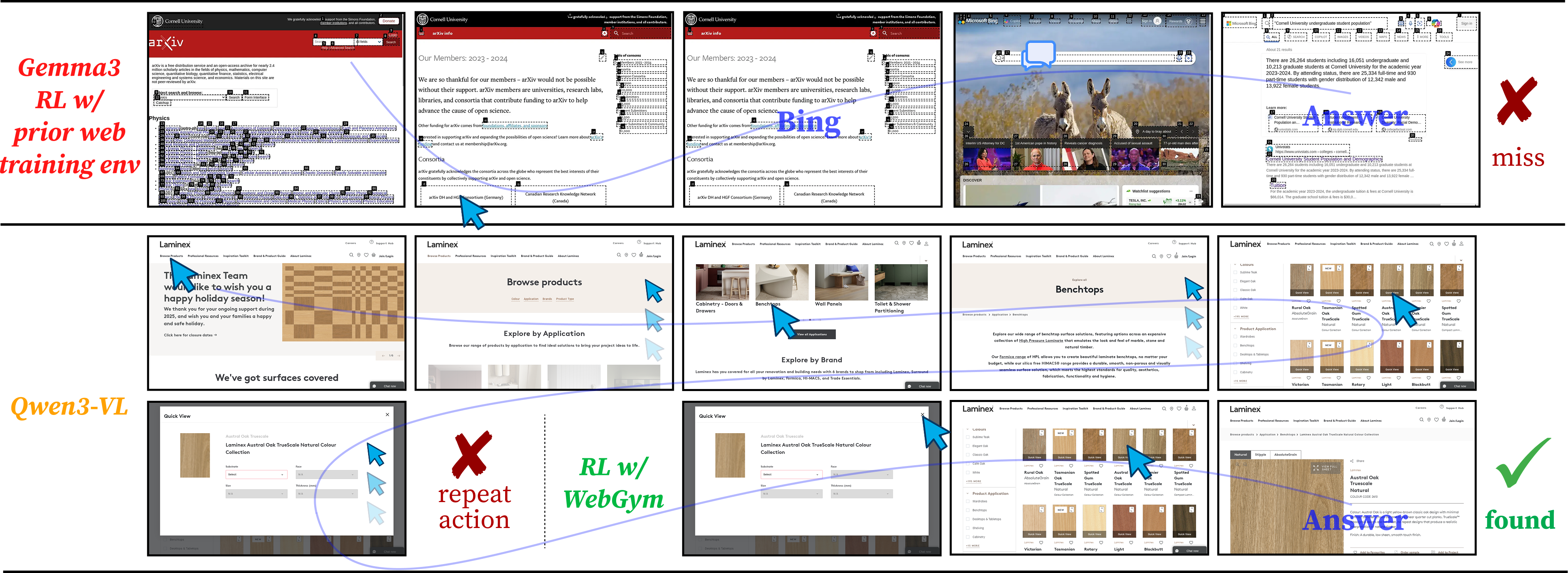}
\captionof{figure}{\footnotesize \textbf{\textit{Example rollouts from visual web agents trained on different training environments.}} Tasks in prior large-scale training setups were relatively simple, e.g., \textcolor{red}{Test-Time-Interaction} (TTI;~\citet{test-time-interaction}, \textit{Row 1}), resulting in failures of trained agents on many held-out tasks (task shown: \textit{From ArXiv, access the website of the university that maintains and manages ArXiv. How many undergraduate students are currently at the university?}, where the agent fails to access the Cornell website). We build~\textcolor{orange}{\ours{}} (\textit{Row 2, bottom}), a significantly larger training environment supporting harder and more diverse tasks, totaling nearly 300k tasks (3$\times$ the size of TTI), and train an agent via online reinforcement learning to acquire more generalizable skills (task shown: \textit{Find the product code for the `Austral Oak TrueScale' laminate benchtop design on laminex.com.au.}, where the base model (untrained) agent gets stuck repeating its behavior (\textcolor{red}{\textit{repeat action}}). In contrast, the agent trained with \ours{} (\textcolor{mygreen}{RL w/ \ours{}}) successfully solves the task).}
\label{fig:teaser}
\end{center}
% \vspace{0.5em}

% Abstract box
\begin{tcolorbox}[
    colback=aifrontiersabstractbg,
    colframe=aifrontiersabstractframe,
    boxrule=0.5pt,
  breakable, enforce breakable,
  shrink break goal=0.5\baselineskip,
    enhanced,
    arc=8pt,
    left=8pt,
    right=8pt,
    top=6pt,
    bottom=6pt
  ]
We present \emph{WebGym}, the largest-to-date open-source environment for training realistic visual web agents. Real websites are non-stationary and diverse, making artificial or small-scale task sets insufficient for robust policy learning. WebGym contains nearly 300,000 tasks with rubric-based evaluations across diverse, real-world websites and difficulty levels. We train agents with a simple reinforcement learning (RL) recipe, which trains on the agent's own interaction traces (\textit{rollouts}), using task rewards as feedback to guide learning. To enable scaling RL, we speed up sampling of trajectories in \ours{} by developing a high-throughput asynchronous rollout system, designed specifically for web agents. Our system achieves a 4-5x rollout speedup compared to na\"ive implementations. Second, we scale the task set breadth, depth, and size, which results in continued performance improvement. Fine-tuning a strong base vision-language model, \textit{Qwen-3-VL-8B-Instruct}, on \ours{} results in an improvement in success rate on an out-of-distribution test set from 26.2\% to 42.9\%, significantly outperforming agents based on proprietary models such as GPT-4o and GPT-5-Thinking that achieve 27.1\% and 29.8\%, respectively. This improvement is substantial because our test set consists only of tasks on websites never seen during training, unlike many other prior works on training visual web agents.
\end{tcolorbox}

\vspace{-0.5cm}
\section{Introduction}
\label{sec:intro}

Vision-language models (VLMs) have recently demonstrated remarkable performance on agentic tasks on the web. These tasks often resemble those that a human user would aim to perform on the web, such as \emph{``Compare the prices and specs between AirPod 3 and AirPod 2 on Apple.com"}. To solve such tasks, an agent will perform a series of actions, such as navigating to the relevant website, searching for both products, and finally, deciding on and submitting an answer to complete the task.
Some web agents rely purely on textual information and try to solve tasks by simply utilizing accessibility trees~\cite{webarena} as the observation for the agent, while we focus on visual web agents, where agents observe screenshots, much like humans. This choice is based on the premise that web interaction semantics are defined by the rendered interface users see, and that reasoning over visual affordances enables more robust generalization across diverse and dynamically generated web pages~\citep{visual-web-arena, pae-webvoyager, test-time-interaction}.

There is a growing body of work on evaluating visual web agents by using benchmarks that involve either artificial or real-world websites. Benchmarks with artificial websites typically provide source code for hosting simulated environments that resemble real websites, such as REAL Bench~\citep{real-bench} and VisualWebArena~\citep{visual-web-arena}. In contrast, realistic benchmarks such as GAIA-Web~\citep{gaia}, BrowseComp~\citep{browsecomp}, and InSTA-v3~\citep{insta-v3} use tasks based on live websites. These ``live-website'' benchmarks tend to be more challenging, as real websites are inherently non-stationary~\citep{digirl, digi-q}. For example, each time an agent visits the homepage of an e-commerce site, the set of displayed products may differ, and the same sequence of past actions could yield very different outcomes.

Recent work has demonstrated that vision-language models can be adapted into web agents via supervised fine-tuning and task-specific training pipelines~\citep{glm-4.1v}. However, in contrast to the rapid progress of online reinforcement learning (RL) in text-only domains such as software engineering and mathematical reasoning~\citep{swe-gym,e3}, scaling RL for visual web agents remains substantially more challenging. Unlike text-based settings, where rollouts are fast and rewards are easily verifiable, visual web tasks require reasoning over fine-grained rendered interfaces and often lack unambiguous success signals~\citep{vision-r1,sarch2025grounded,online-mind2web}, making data collection slow and inefficient.  These challenges have inhibited scaling up post-training methods for visual web agents, particularly for diverse and long-horizon tasks, beyond current short-horizon settings~\citep{pae-webvoyager}.

Thus, to train advanced visual web agents, we first need a strong \textbf{\emph{task}} set that includes problems which are challenging for existing models, and that span diverse domains and goals to encourage generalizable policy learning. \emph{A good task set should be able to support scaling performance by scaling these dimensions.} These tasks must also be paired with clear \textbf{\emph{evaluation}} protocols that provide meaningful learning signals. Such a task set would need to be substantially larger and more varied than existing ones~\citep{pae-webvoyager}. In addition, utilizing a large task set requires a high \textbf{\emph{training speed}}, with rollout efficiency being the main bottleneck. Rollouts in visual web environments typically take far longer than the learning updates themselves, largely due to the high computational overhead of simulating web browsers in current RL pipelines.

To advance scaling of training recipes for visual web agents, we introduce \ours{}, a training environment featuring \textbf{the largest open-source task set and the fastest rollout system to date} (to our knowledge) for training web agents. Specifically, \ours{} provides \textbf{(1)} a task set with nearly \textit{300,000} tasks spanning a wide range of difficulties, domains, and evaluation criteria, enabling agents to generalize across diverse and challenging scenarios, and \textbf{(2)} an efficient asynchronous rollout system capable of collecting \textit{1,800 trajectories} with an average of \textit{13.2 steps} in just \textit{30 minutes} using \textit{128 CPUs} and \textit{24 NVIDIA H100 GPUs}. This achieves a \textit{4-5$\times$} speedup compared to traditional synchronous rollout systems. We study three dimensions of scaling the task set in \ours{}: \emph{task set breadth, depth and size}, and observe significant improvements over each dimension with the task set introduced in \ours{}. With these improvements, we demonstrate that even a simple REINFORCE-like training algorithm can improve the performance of Qwen3-VL-Instruct-8B, a strong vision-language model to \emph{state-of-the-art} performance on its scale on web agent benchmarks, across a wide spectrum of tasks, ranging from easy to long-horizon in strongly out-of-domain settings.

\section{Related Work}
\label{sec:related-work}

\textbf{Visual web agents for realistic tasks.}
This work focuses on visual web agents designed to solve real-world tasks using vision-language models (VLMs). Such agents emulate how humans interact with browsers on websites: they take website screenshots and metadata as input, and generate interactive actions supported by the browser (e.g., clicking or typing) to make progress toward a given goal. This process repeats until the agent outputs an action indicating task completion, resulting in multi-step trajectories. Prior to this work, both general-purpose VLMs and web-specific visual agents have shown promising capabilities on this problem of training visual web agents. \textit{General-purpose VLMs} can perform reasonable grounding and reasoning on web tasks when combined with prompt engineering or light post-training on web-related data to improve general agent behavior, as demonstrated by models such as Gemma3~\citep{gemma3}, GLM-4.1v~\citep{glm-4.1v}, and Qwen3-VL~\citep{qwen3-vl}. In contrast, \textit{web-specific visual agents} are models that are heavily post-trained on web/web-adjacent tasks, like UI-TARS~\citep{ui-tars}.

\textbf{Evaluation benchmarks and training environments for realistic web agents.}
Recent work has made substantial progress in developing robust \emph{evaluation benchmarks} for realistic web-browsing agents, while \emph{training environments} remain relatively scarce. Existing benchmarks typically feature small, curated task sets with evaluation hints to assess specific reasoning or interaction abilities. Representative visual web benchmarks include WebVoyager~\citep{webvoyager}, Mind2Web-Live~\citep{mind2web-live}, and Mind2Web-2~\citep{mind2web-2}, alongside broader but non-visual ones such as AgentSynth~\citep{agentsynth}, InSTA~\citep{insta-v3}, BrowseComp~\citep{browsecomp}, TravelPlanner~\citep{travelplanner}, DeepShop~\citep{deepshop}, and GAIA~\citep{gaia}. A \emph{training environment}, in contrast, must provide not only large-scale task sets but also efficient rollout and training systems. Efforts like WebRL~\citep{webrl} and the Proposer-Agent-Evaluator (PAE) framework~\citep{pae-webvoyager} have taken steps in this direction, but their task sets remain limited in diversity and hardness\footnote{In this paper, ``difficulty'' or ``difficulty level'' specifically means the difficulty levels we propose as an attribute of the task set we propose, while we use the word ``hardness'' for indicating whether a given task is difficult or not.}, and their rollout systems are not optimized for speed.
\ours{} addresses these issues by expanding task diversity via rubric-derived fact groups and controlled decomposition into atomic and compositional variants (while limiting additional synthesis for sources like PAE-WebVoyager and AgentSynth to avoid near-duplicate synthetic goals), and by using a fully asynchronous rollout system that accelerates data collection and enables scalable reinforcement learning for web agents.

\textbf{Scaling online reinforcement learning for training agents.} Prior work has improved the performance of LLMs and VLMs by scaling the size and diversity of supervised fine-tuning (SFT) datasets, such as Super Natural-Instruct~\citep{super-natural-instruct} and OpenThoughts~\citep{open-thoughts}. More recently, studies have shown that increasing the number of training trajectories through online RL can further boost agent capabilities, even with algorithms like REINFORCE~\citep{original-reinforce}, across software engineering~\citep{swe-gym}, math reasoning~\citep{e3}, deep research~\citep{deepresearcher-rl, search-r1, paper-search-agent}, and general reasoning. \ours{} builds on these insights by constructing a large, diverse task set spanning multiple domains and difficulty levels, and training agents via on-policy RL from collected rollouts.

\section{Tasks in \ours{} and Evaluation Protocol} \label{sec:task}

Training with online RL requires a large number of model-generated rollouts guided by reliable reward signals on a family of tasks (``environments'' in traditional multi-task RL nomenclature) that we curate. To learn generalizable policies with such a procedure, the task set must \textbf{(1)} span diverse domains and websites, \textbf{(2)} cover a range of ``difficulties'' from atomic to compositional tasks, and \textbf{(3)} include verifiable evaluators that translate outcomes into meaningful learning signals. Existing web agent benchmarks, while effective for evaluation, are typically too small and static for large-scale training. Existing training tasks are either non-diverse~\citep{pae-webvoyager} and lack a wide range of task difficulties~\citep{insta-v3}, or do not come paired with good task-level evaluation criteria, rubrics, or guidance for enabling robust policy evaluation~\citep{insta-v3, pae-webvoyager, agentsynth, browsecomp, gaia}.
We describe how we procedurally construct a task set that satisfies these desiderata, with assistance from LLMs.

We start from a seed collection of high-quality tasks drawn from existing benchmarks and training environments (\S\ref{sec:task/source}). We systematically annotate these tasks with evaluation criteria (evaluation references) and then decompose them into subtasks (\S\ref{sec:task/algorithm}) to achieve both \textit{breadth} (coverage across many websites and domains) and \textit{depth} (multi-criterion compositional tasks) (\S\ref{sec:task/stats}). The construction is fully done with the GPT-4o model and all prompts are provided in~\S\ref{app:prompt} to ensure reproducibility and enable scaling up further. We then detail how we split the task set into train and test subsets (\S\ref{sec:task/split}) and present our evaluation protocol (\S\ref{sec:task/eval}).

\begin{figure}[t!]
\centering
\begin{minipage}[t]{0.48\textwidth}
\vspace{0pt}
\centering
\resizebox{\linewidth}{!}{
\begin{tabular}{lccc}
\toprule
\textbf{Source Task Set} & \textbf{Difficulties} & \textbf{\# Websites} & \textbf{\# Tasks} \\
\midrule
InSTA-v3 \citep{insta-v3} & \xmark & 146,348 & 146,441 \\
PAE-WebVoyager \citep{pae-webvoyager} & \xmark & 13 & 128,499 \\
AgentSynth-Web \citep{agentsynth} & \cmark & 328 & 2,086 \\
BrowseComp \citep{browsecomp} & \xmark & 7 & 1,266 \\
TravelPlanner \citep{travelplanner} & \xmark & 1 & 1,225 \\
Mind2Web-Live \citep{mind2web-live} & \xmark & 76 & 542 \\
Online Mind2Web \citep{online-mind2web} & \xmark & 139 & 300 \\
DeepShop \citep{deepshop} & \xmark & 1 & 150 \\
Mind2Web-2 \citep{mind2web-2} & \xmark & 44 & 130 \\
GAIA-Web \citep{gaia} & \cmark & 1 & 87 \\
\midrule
\ours{} (Ours) & \textbf{\textcolor{red}{\cmark}} & \textbf{\textcolor{red}{127,645}} & \textbf{\textcolor{red}{292,092}} \\
\bottomrule
\end{tabular}
}
\captionof{table}{\textbf{Task sourcing for \ours{}.} We aggregate web agent task sets from 10 widely-used benchmarks and environments to seed our procedural construction. We report original statistics (\#difficulties, \#websites, \#tasks) for each seed set used to construct the \ours{} task set. Some task sets are marked with only 1 website because the original task sets do not provide task-specific website.}
\label{tab:swe-gym}
\end{minipage}
\hfill
\begin{minipage}[t]{0.48\textwidth}
\definecolor{darkgreen}{rgb}{0, 0.5, 0}
\footnotesize
\hrule
\vspace{6pt}
\begin{algorithmic}[1]
\State Let $C$ be the task set with $N$ tasks, each having task description, difficulty, domain, website, and evaluator rubric fields.
\State Gather all tasks from datasets included in \ours{}.
\For{each task in $C$}
\State Infer website and domain from task description.
\State Generate evaluator rubric as fact groups using LLM, where each group contains one or more facts.
\State Set difficulty as total number of facts.
\If{at least $2$ groups exist \textbf{and} at least one group has $3$ or more facts}
\For{each proper subset of groups containing at least one large group}
\State Create decomposed task from subset using LLM, set its rubric and difficulty, then add to $C$.
\EndFor
\EndIf
\EndFor
\end{algorithmic}
\vspace{6pt}
\hrule
\vspace{3pt}
\captionof{algorithm}{\footnotesize Task Set Construction Procedure of \ours{}.}
\label{algo:task/task-set-construction}
\vspace{-0.5cm}
\end{minipage}
\end{figure}

\subsection{Obtaining a Seed Set of Tasks} \label{sec:task/source} 

To obtain a set of seed tasks, we collect \textbf{10} of the most recent and highly utilized task sets in the open-source community, as listed in~\Cref{tab:swe-gym}. These tasks come from both existing benchmarks and existing training sets. Importantly, we do not designate any entire benchmark as ``held out'', because many benchmarks share websites, domains, or task patterns with one another, making train-test separation at the level of existing benchmarks unreliable. Instead, \ours{} constructs its own train-test split at the task and website level to ensure strict separation without discarding large amounts of useful data. Specifically, our tasks come from \textbf{a)} InSTA-v3~\citep{insta-v3}, which provides wide coverage of websites and domains; \textbf{b)} PAE-WebVoyager~\citep{pae-webvoyager} and AgentSynth~\citep{agentsynth}, which provide synthetic tasks focused on narrow domains; \textbf{c)} BrowseComp~\citep{browsecomp} and GAIA-Web~\citep{gaia}, which include difficult tasks with concrete, verifiable answers; \textbf{d)} TravelPlanner~\citep{travelplanner} and DeepShop~\citep{deepshop}, which focus on domain-specific browsing problems; and \textbf{e)} Mind2Web-Live~\citep{mind2web-live}, Online Mind2Web~\citep{online-mind2web}, and Mind2Web-2~\citep{mind2web-2}, which contain human-verified high-quality tasks that are indeed solvable. We next describe how we construct our full task set.

\begin{figure}[t!]
\centering
\includegraphics[width=\linewidth]{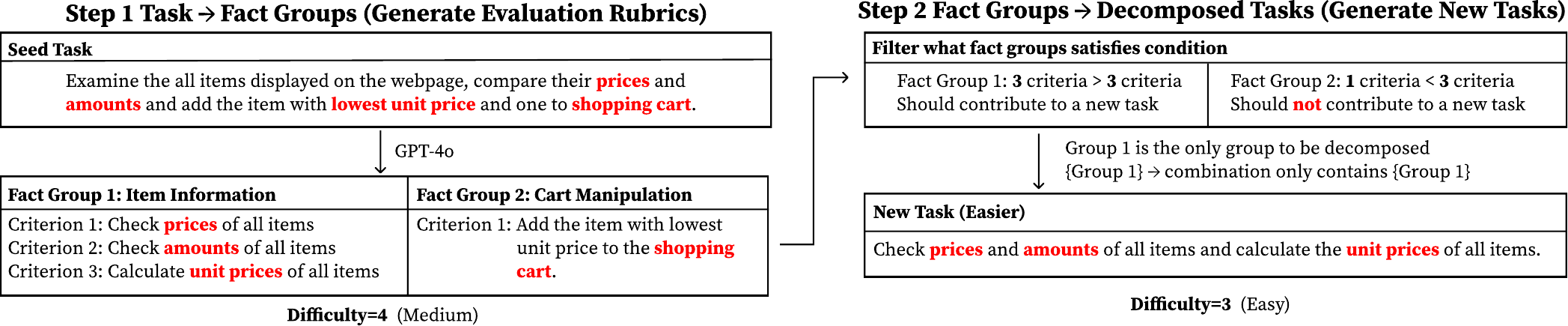}
\caption{\textbf{Task decomposition system.} \ours{} decomposes tasks by generating valid combinations of fact groups from the original task's rubric. Decomposition requires $\geq$2 groups with at least one ``large'' group ($\geq$3 facts). Each valid combination (excluding the full set) produces a new task with lower difficulty while maintaining consistency with the original objectives.}
\label{fig:env/decompose_task_example}
\end{figure}

\subsection{Task Set Construction Procedure} \label{sec:task/algorithm}
Recent research has demonstrated strong improvements when constructing curriculum on tasks with similar goals but different hardness~\citep{test-time-interaction, e3}. In parallel, other efforts show that developing evaluation rubrics can significantly improve evaluation accuracy~\citep{task-specific-rubric, rubric-anchors, rubrics-as-rewards}. We combine these ideas by generating evaluation rubrics and using them to construct decomposed tasks. We elaborate our task construction algorithm in~\Cref{algo:task/task-set-construction} and an example in~\Cref{fig:env/decompose_task_example}. All constructions are generated using GPT-4o.

We begin by sourcing and merging all tasks specified in~\S\ref{sec:task/source}. For benchmarks that do not specify a concrete website for a task, we prompt GPT-4o to infer an appropriate website. Next, we use GPT-4o to generate an evaluation rubric structured as fact groups, where each group contains one or more evaluation facts. We define the \emph{difficulty} or \emph{difficulty level} of a task as the total number of facts across all groups in its evaluation rubric. The prompt we use to generate the evaluation rubric is shown in~\S\ref{app:prompt/task/criteria}.

With these fact groups, we can procedurally generate decomposed tasks through combinations. Task decomposition occurs only when two conditions are met: (1) the original task has at least 2 fact groups, and (2) at least one group is ``large'' (i.e., it contains 3 or more facts). When these conditions are satisfied, we generate new tasks by selecting proper subsets of the fact groups, where each subset must contain at least one ``large'' group to ensure meaningful task complexity. For example, as illustrated in~\Cref{fig:env/decompose_task_example}, the seed task ``\textit{Examine all the items displayed on the webpage, compare their prices and 
amounts and add the item with lowest unit price and one to shopping cart}'' contains 2 fact groups, with one fact group containing 3 criterion while the other fact group containing only 1. In this case, only the first fact group will participate in the decomposition, and thus our algorithm generates only 1 new task with the only ``large'' group.

Take a harder task for example, such as: ``\textit{Find an eligible upcoming piano concert in the US or Canada by a Chopin Competition prize winner, providing the pianist's name, competition year/prize, YouTube link to their final performance, and the concert's date, city, venue, and event page link.}'' We decompose this into three fact groups: G1 (concert eligibility) with 2 facts (timing and location constraints), G2 (pianist details) with 3 facts (name, competition prize, YouTube link), and G3 (concert details) with 4 facts (date, city, venue, event link), yielding a total difficulty of 9. Since the decomposition conditions are met (3 groups $\geq$ 2, and G2 and G3 are ``large'' with $\geq$3 facts each), we generate valid subset combinations: \{G2\} (d=3), \{G3\} (d=4), \{G1,G2\} (d=5), \{G1,G3\} (d=6), and \{G2,G3\} (d=7). The combination \{G1\} alone is invalid since G1 has only 2 facts (no large group), and \{G1,G2,G3\} is excluded as it equals the original task. Example rubrics can be found in~\S\ref{app:qual/rubric}. We show the prompt we use to generate the decomposed tasks in~\S\ref{app:prompt/task/task}.

As a result, tasks generated by \ours{} (1) avoid being trivial by restricting to sufficiently large groups, (2) are strictly easier than the original task, yielding denser reward signals, and (3) are guaranteed to be well-defined whenever the original task is well-defined. In contrast, PAE~\citep{pae-webvoyager} generates synthetic tasks that may be ill-posed or impossible to solve. While \ours{} is related to the approach in AgentSynth~\citep{agentsynth}, it does not require executing agent rollouts to generate new tasks making it cheaper.

\subsection{Train-Test Split} \label{sec:task/split}

Having expanded the source tasks into a substantially larger task set, we next describe the train-test split. As we're primarily interested in imbuing agents with broadly generalizable skills and capabilities, we construct an out-of-distribution (OOD) task set with 1,167 tasks where each OOD task originates from a distinct website. We remove all tasks in the training set associated with the same websites as those in the test set. This guarantees that all OOD tasks come from \textit{entirely unseen websites} relative to the training set.

\subsection{Statistics: Size, Breadth and Depth} \label{sec:task/stats}

\begin{figure*}[!t]
  \centering
  \includegraphics[width=0.48\linewidth]{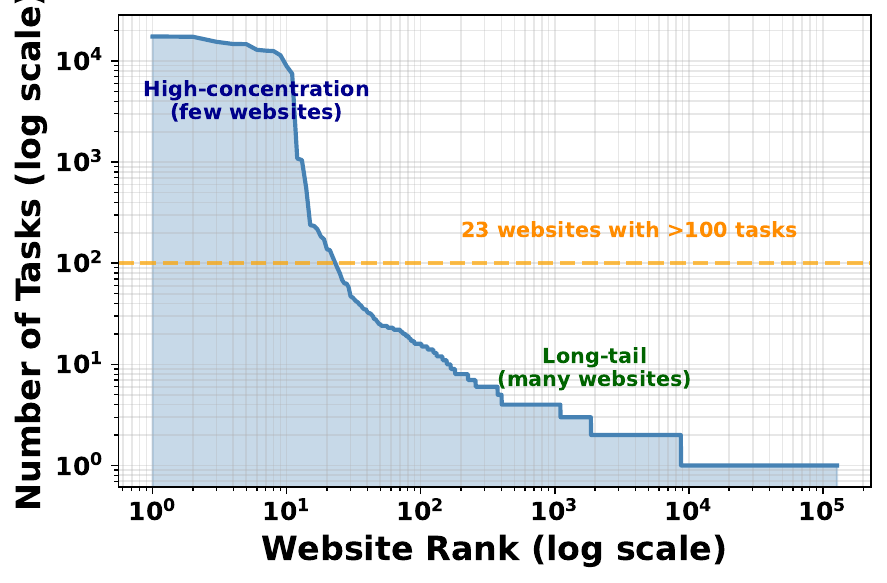}
  \hfill
  \includegraphics[width=0.48\linewidth]{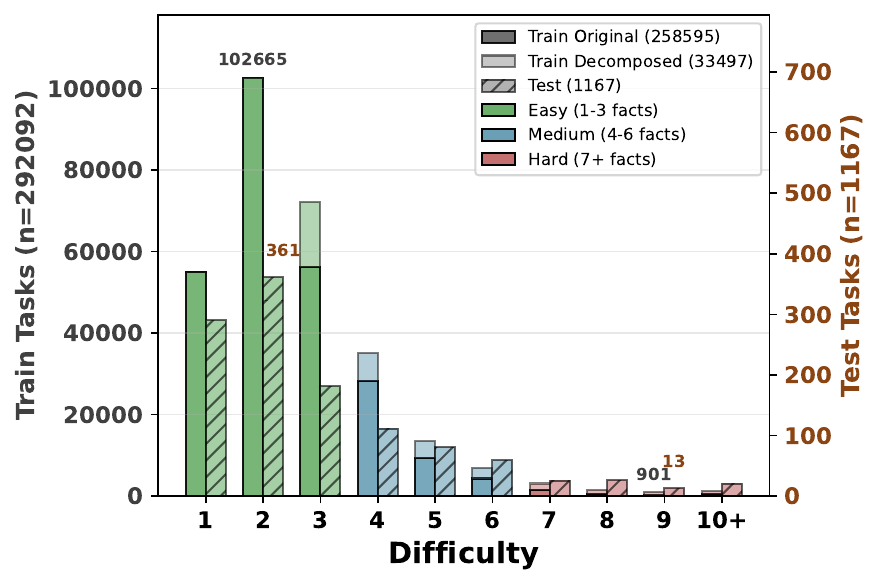}
  \vspace{-0.2cm}
  \caption{\textbf{Statistics of the \ours{} \textit{training} task set.} \textit{Left:} website distribution as a function of the sorted index (from more tasks to less) of website. \textit{Right:} difficulty distribution of the train and test task sets. The transparent bars over the original bar mean decomposed tasks, and the slashed bars means test set.]}
  \label{fig:task/stats-top}
  \vspace{-0.1cm}
\end{figure*}

\begin{figure}[t]
\centering
\begin{subfigure}[b]{0.40\linewidth}
    \centering
    \includegraphics[width=\linewidth]{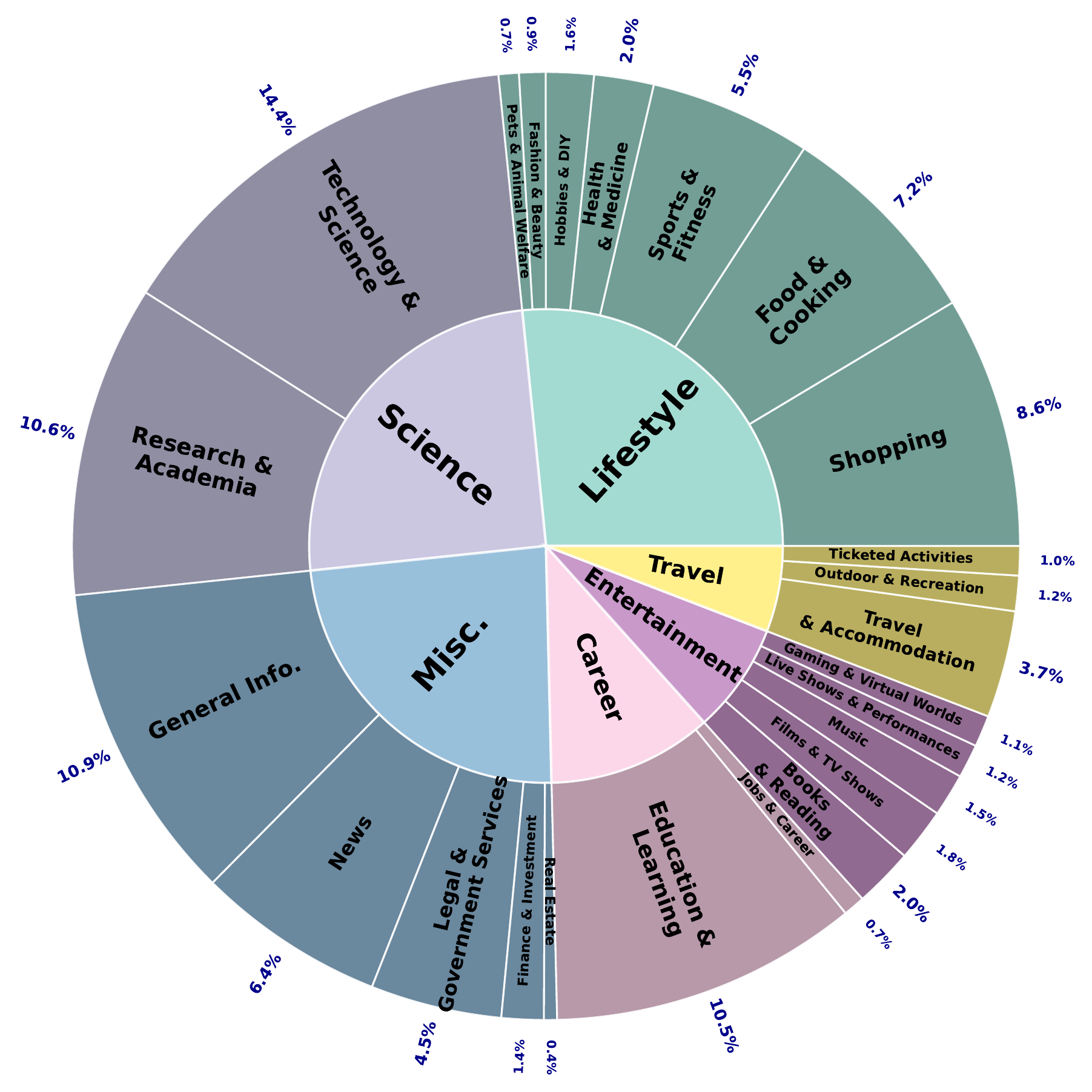}
\end{subfigure}
\hfill
\begin{subfigure}[b]{0.52\linewidth}
    \centering
    \includegraphics[width=\linewidth]{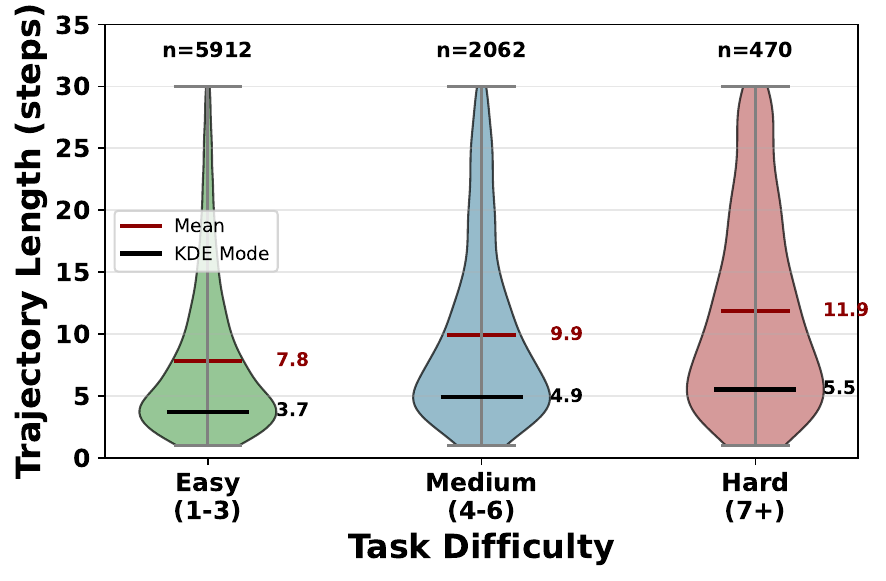}
\end{subfigure}
\caption{\textbf{Analysis of the \ours{} task set.} \textit{Left:} distribution of tasks across domains according to the Mind2Web-2 taxonomy~\citep{mind2web-2}. \textit{Right:} Distribution of trajectory lengths by task difficulty for \textit{answered} trajectories over multiple iterations. For trajectories working on medium- and hard-difficulty tasks that exceeds 30 tasks during evaluation, we filter in only trajectories under 30 tasks to make the comparison fair. The violin plots show that the probability density of trajectory lengths, with the KDE mode (black line, KDE mode is the peak of the smoothed probability density curve, where the violin is widest) indicating the most likely length and mean (red) indicating the average.}
\label{fig:task/stats-bottom}
\end{figure}

After the train-test split, the \ours{} training set contains 292,092 tasks spanning 127,645 websites. The top 20 websites account for about half of all tasks (\Cref{fig:task/stats-top}, \textit{left}), most of which are from the PAE-WebVoyager synthetic task set, while there are a significant number of websites with few tasks that build the diversity up, which is majorly composed of other source task sets. This reflects an \emph{intentional} balance: many long-tail websites ensure cross-domain diversity, while frequent websites provide sufficient repetition for effective RL training. The task difficulty distribution (\Cref{fig:task/stats-top}, \textit{right}) shows that around 80\% of the tasks are Easy tasks (difficulty 1-3), which is intentional to encourage learning generalizable basic skills while we focus on more specific objectives at higher difficulty levels. To verify topical coverage beyond high-frequency sites, we compute domain diversity using the Mind2Web-2 taxonomy (6 domains, 24 subdomains) with GPT-4o. \Cref{fig:task/stats-bottom} (left) illustrates broad domain coverage, with rollouts from different domains to promote diverse training samples. The trajectory length analysis (\Cref{fig:task/stats-bottom}, right) reveals a clear correlation between task difficulty and the number of steps required: easy tasks (difficulty 1-3) require 7.8 steps, medium tasks (difficulty 4-6) require 9.9 steps, and hard tasks (difficulty 7+) demand 11.9 steps on average. The KDE mode (which measures most likely trajectory length) also increases with difficulty, from 3.7 to 4.9 to 5.5 steps. This trend confirms that the tasks classified to be "harder" in \ours{} do actually involve more complex multi-step action and navigation before the agent can provide an answer.

\subsection{Evaluator and Reward Function Design} \label{sec:task/eval}

\begin{wrapfigure}{r}{0.60\textwidth}
  \centering
    \includegraphics[width=0.99\linewidth]{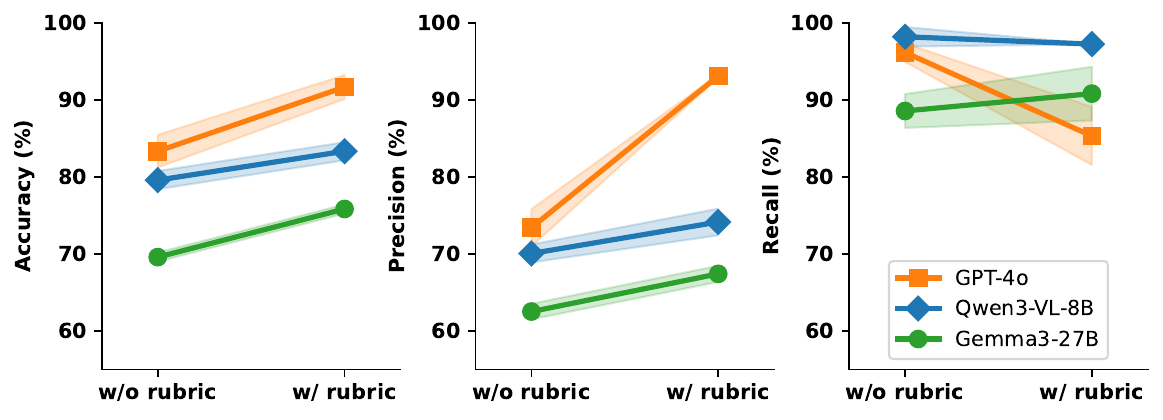}
    \caption{\textbf{Agreement between automated evaluators and human judgment.} Rubric-based evaluation (with explicit criteria) consistently improves agreement over task-only evaluation, yielding higher accuracy and precision across LLM-based evaluators. Among the evaluators, GPT-4o shows the largest shift after adding the rubric: precision increases the most while recall drops, indicating that the rubric makes GPT-4o apply stricter, more conservative pass criteria.}
    \label{fig:env/rubric-comparison}
\end{wrapfigure}

Evaluating long-horizon visual agents is challenging because the produced trajectories often make partial progress, end up at visually similar but incorrect pages, and generate answers that may appear correct unless verified against grounded evidence. Unlike math reasoning or coding, ``reference'' answers of any sort are rare to obtain for most task instructions because only a small minority of benchmarks provide them (for example BrowseComp and GAIA-Web~\citep{browsecomp, gaia}) and the notion of a reference answer does not exist in many cases (e.g., tasks that do not involve question answering). This difference reflects a fundamental challenge in designing open-ended web tasks with a binary answer. Thus, evaluation cannot rely on direct matching to ground-truth outputs and must instead interpret the trajectory in context. To this end, we mainly focus on LLM-judged evaluation, while reference answers, when available in the source benchmark, overrides the judge. 

For faithful evaluation, recent research has adopted a criterion proposal approach that generates task-specific criteria to guide evaluation~\citep{k2, chen2025rm}. However, we posit that tasks usually come with structures, so we propose to generate rubric with structures such as fact groups. As discussed in~\Cref{sec:task/algorithm}, in \ours{}, each task is paired with a rubric of several fact groups, where each group contains one or more criteria that guides evaluation reducing false positives from ambiguity about task success. 

With the task-specific rubric, a trajectory receives reward only when all criteria are satisfied, producing a binary signal consistent across domains and difficulty levels. Since not all screenshots are informative, the judge focuses on evidence-bearing pages, obtained via a high-recall keypoint selection process inspired by Online Mind2Web~\citep{online-mind2web} that retains relevant observations while filtering out distractors such as detours and ads. In practice, we use GPT-4o for both keypoint selection (prompt shown in~\S\ref{app:prompt/eval/keypoint} and per-criterion evaluation (prompt shown in ~\S\ref{app:prompt/eval/evaluation}). The evaluator proposes keypoint screenshots from the task description, iterates through trajectory steps, and issues binary judgments for each criterion using the selected keypoints and final answer. 

We validate the evaluator against human annotations on 80 collected trajectories sampled uniformly across difficulty levels (10 trajectories per level from difficulties 1-6, and around 5 trajectories per level from difficulties 7-10), where nearly half of the trajectories in each difficulty are marked as correct by the \ours{} evaluator to balance the set, and find that rubric-guided evaluation improves agreement over task-only judging,  which increases accuracy and precision for all evaluator models being tested (GPT-4o, Qwen3-VL-8B-Instruct, and Gemma3-27B-it). We observe minor regressions in the recall metric for the subset of stronger evaluator models (Qwen3-VL-8B-Instruct and especially GPT-4o), where some rubrics tend to be overly strict (\Cref{fig:env/rubric-comparison}). 
Note that this regression is almost impossible to eliminate when using an LLM-based rubric proposer: depending on the prompt used to instruct the model, it may generate stricter rubrics, occasionally resulting in over-strict evaluations. Examples of evaluations with and without criteria are shown in~\S\ref{app:qual/eval}. This trade-off is ultimately beneficial for RL training: we exchange a slight reduction in sample efficiency for more precise but conservative learning signals, but at the same time, this helps prevent error accumulation in long-horizon settings.
\section{Rollout System in \ours{}} \label{sec:env}

The primary bottleneck in scaling up RL training for web agents lies in the slow speed of rollout generation, particularly in multi-step RL settings. This challenge becomes more severe when simulation resources, such as CPUs, are limited. To make \ours{} practical for large-scale agent training, we address this issue by replacing synchronized, batch-style rollout collection with an asynchronous system that \textit{keeps both CPUs and GPUs fully utilized} throughout multi-step rollout generation. We show in  Figure~\ref{fig:env/async_rollout} how tasks are queued into a process pool, how stateful browser sessions are isolated for each rollout, and how new tasks immediately begin inference once resources become available, eliminating the need for global synchronization barriers. We now describe this system and benchmark its performance and throughput. 

To our knowledge, the \ours{} provides the \textbf{the first, open-source rollout system} optimized for asynchronous multi-turn simulations for web agent tasks. In contrast to work on RL infrastructure for LLMs, \ours{} focuses on an orthogonal dimension: speeding up rollout generation up by introducing a highly efficient simulation server for web browsing so that the rollout pipeline \textit{does not synchronize between any step or episode}. This server can be easily integrated with any RL framework that focuses on system-level speedup like optimizing the weights save/load overhead and algorithmic speedup like introducing episode/step/token-level off-policy RL. Examples of these RL frameworks include veRL~\citep{verl}, veRL-agent~\citep{verl-agent}, PipelineRL~\citep{pipeline-rl}, SkyRL~\citep{skyrl-agent}, and AReaL~\citep{areal}.

We will use the simplest implementation of the RL loop in our experiments, where rollout and training happen synchronously. We instead focus on optimizing rollout generation as discussed above. Since we only optimize rollout generation, in this section, our benchmarking numbers are based on inference-only experiments, and no training is involved.

\subsection{Strategy and System Architecture}

\begin{figure*}[t]
\small
  \centering
  \begin{minipage}[t]{0.49\textwidth}
    \centering
    \includegraphics[width=\linewidth]{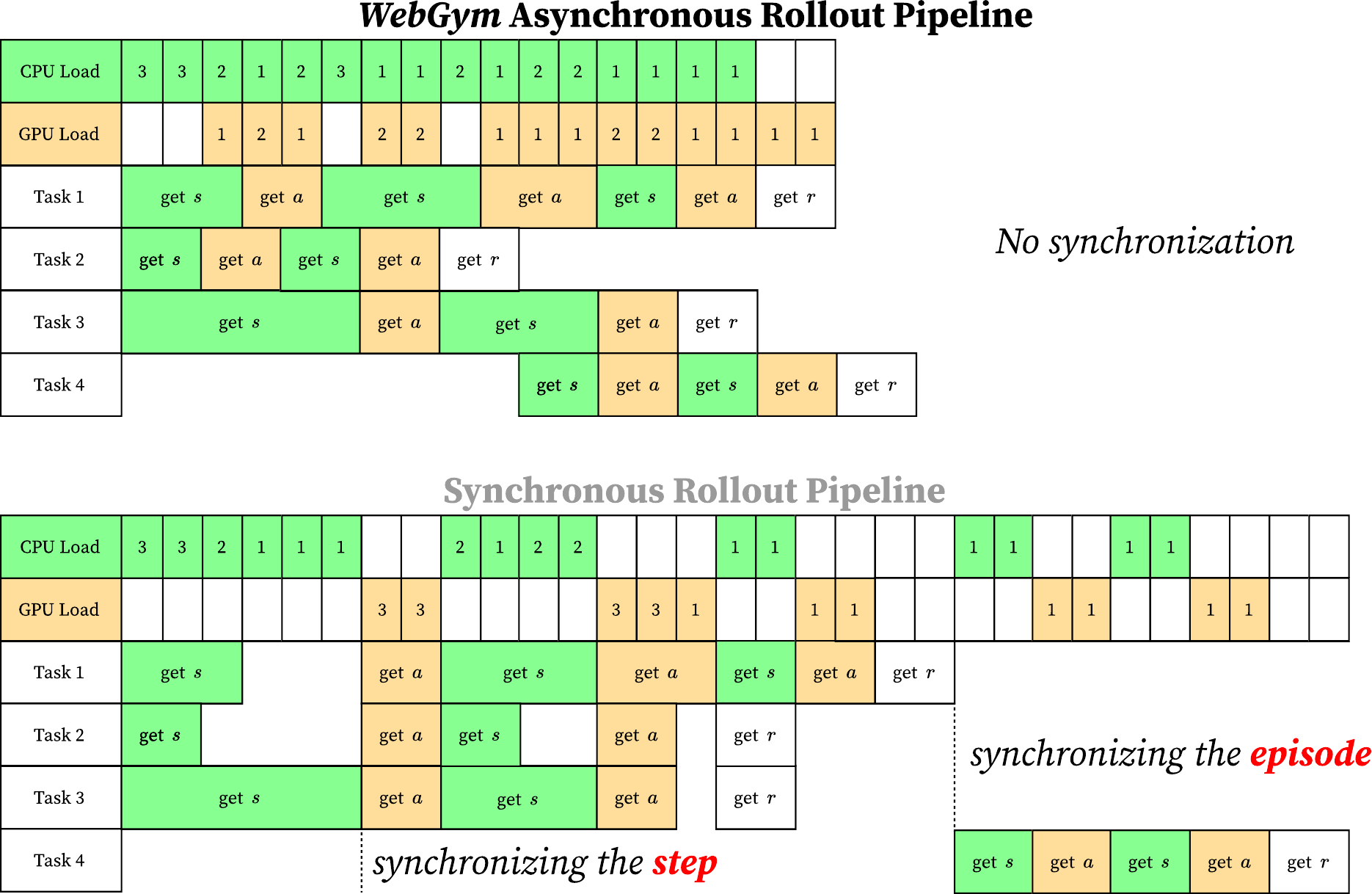}
  \end{minipage}\hfill
  \begin{minipage}[t]{0.49\textwidth}
    \centering
    \includegraphics[width=\linewidth]{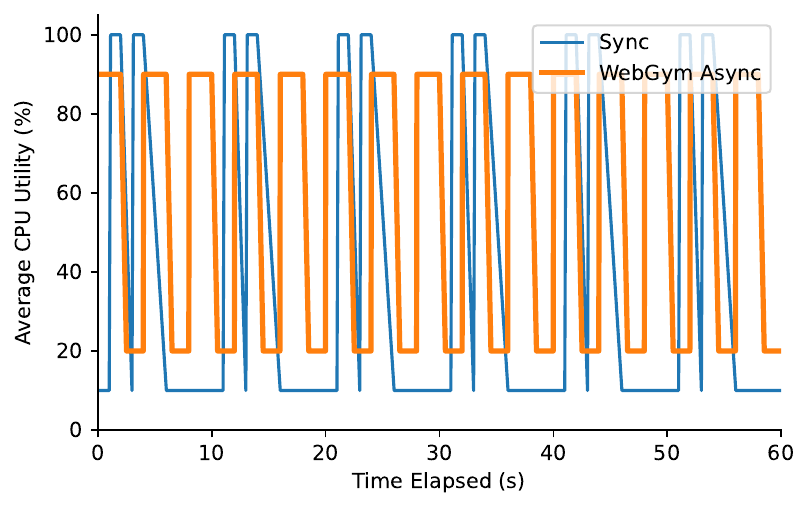}
  \end{minipage}
  \caption{\textbf{Asynchrony eliminates burst-idle behavior in web rollouts.}
  \textit{\textbf{Left:}} \ours{} implements an \textbf{\textit{asynchronous rollout system}} that (1) shortens single-trajectory collection duration by isolating rollout processes and (2) allows new rollout processes to join early by replacing batched inference with a process pool; the example shown is a toy case with 3 available environment buckets but 4 tasks waiting to roll out.
  \textit{\textbf{Right:}} CPU utility trace over a sampled rollout period, comparing the synchronized rollout framework to \ours{}: synchronized barriers induce spiky ``all-busy then all-idle'' utilization, whereas \ours{} streams work as soon as observations are ready, smoothing CPU load and reducing idle gaps while CPU workers await policy actions.}
  \label{fig:env/async_rollout}
\end{figure*}

\begin{figure*}[!htp]
  \centering
  \small
  \begin{minipage}[t]{0.49\textwidth}
    \centering
    \includegraphics[width=\linewidth]{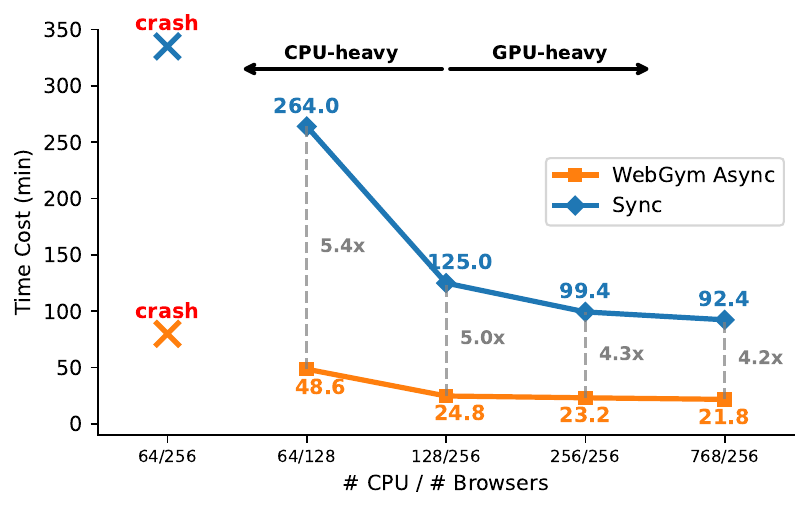}
  \end{minipage}\hfill
  \begin{minipage}[t]{0.49\textwidth}
    \centering
    \includegraphics[width=\linewidth]{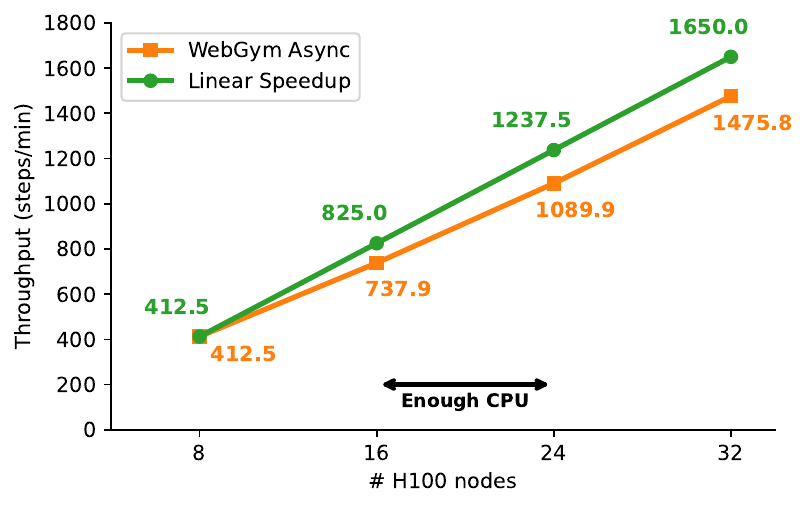}
  \end{minipage}
  \caption{\textbf{Benchmarking speed and throughput of the \ours{} asynchronous rollout framework.}
  \textit{\textbf{Left: the \ours{} framework boosts the rollout speed up significantly with a 4x-5x speedup.}} This figure shows time cost and average CPU utility percentage when reaching all 256 environments running w.r.t. different amounts of CPUs, while using the same amount of GPU resources for running inference on the VLM-based agent. The limit of browsers per CPU can scale to \textbf{\textit{2}} without crashing.
  \textit{\textbf{Right: the \ours{} framework boosts the rollout speed up linearly w.r.t. GPU nodes.}} This figure shows \ours{} throughput w.r.t. different numbers of GPU nodes when enough CPUs are provided on each machine (thus speed is bounded by GPU).}
  \label{fig:env/pressure_test}
\end{figure*}

Na\"ive implementations of rollout systems are \emph{synchronous} in nature, like what was implemented in~\citet{test-time-interaction, digirl, webrl}, where a fixed group of browser sessions is forced to advance at the same pace to perform some number of rollouts in parallel. At each step, every session loads its webpage on the CPUs, the system sends all screenshots from the group together to the GPU for one combined policy call, and then waits until the slowest session completes before any session proceeds; a common variant waits at the end of each full trajectory instead of at every step. These step and episode barriers are problematic in web settings because step times and horizons vary widely across tasks and even across data-collection policies. Faster sessions sit idle while slower ones catch up, and hardware activity arrives in synchronized bursts that saturate either CPUs or GPUs at once, and then drop to idle. We illustrate this ``burst-idle'' behavior in Figure~\ref{fig:env/async_rollout} and then quantify its impact on end-to-end data collection time and scaling behavior in Figure~\ref{fig:env/pressure_test}.

To mitigate the inefficiency from this ``burst-idle'' behavior, our framework adopts a server/client architecture~\citep{server-client-arch}, where the \textit{server} hosts simulation environments and the \textit{client} organizes the rollout loop. The \textbf{server} simulates the browser environments following a master/worker paradigm~\citep{master-worker-paradigm}, with a master node routing API requests to CPU worker nodes that host the actual simulators and execute actions. The master node configuration determines how much parallel work the system can support, making it easy to scale horizontally by adding workers or vertically by increasing per-node resources. The server is entirely CPU-based and requires no GPUs. The \textbf{client} hosts agent instances on the GPU and send requests to the server. To manage the requests sent to the server, we adapt an operation-specific local queue that enables spreading CPU and GPU workload evenly under tight resource settings, with more details in \S\ref{app:env/local-queue}. The client is GPU-heavy and relies little on CPU budgets.

\subsection{Benchmarking our Rollout System} 
We demonstrate the strong capability of the framework under different GPU/CPU resource constraints, comparing \ours{} to a synchronized multi-step rollout strategy, as shown in~\Cref{fig:env/pressure_test} (\textit{left}). We fix the number of GPUs (to 3 nodes of 8 NVIDIA H100 GPUs) and environments for all experiments to collect 1800 trajectories with an \textit{average} (instead of max) of $13.2$ steps thus in total 23,760 steps, then modify only the number of CPUs on the asynchronous rollout server. For all runs, we use CPU clusters where each worker node provides 64 AMD EPYC 7763 CPUs.

We observe a \textbf{4-5$\times$ speedup} with our rollout system compared to the synchronous framework shown in~\Cref{fig:env/async_rollout} (\textit{left}), with greater gains under tighter CPU constraints (\Cref{fig:env/pressure_test}, \textit{left}). Using only 64 CPUs, our system collects 1.8k rollouts in 48.6 minutes, whereas the synchronous system requires 264 minutes. As the number of CPUs increases, the rollout time decreases until around 128 CPUs, beyond which inference throughput plateaus as the three H100 GPU nodes become the primary bottleneck.
In this GPU-limited inference regime, adding more CPUs provides minimal benefit because the GPU inference service becomes saturated, causing requests to queue. This effect appears in \Cref{fig:env/pressure_test} (\textit{left}) as the flattened throughput curve and in \Cref{fig:env/async_rollout} (\textit{right}) as intervals of reduced CPU utilization while workers await GPU responses.

When CPU resources are scarce, the asynchronous design streams observations to the GPUs as soon as they are available, preventing GPU starvation and maintaining high GPU utilization while smoothing CPU load. This contrast between CPU-limited and GPU-limited regimes explains the utilization patterns and measured time-to-trajectory observed in~\Cref{fig:env/pressure_test}. We also benchmark the scalability of \ours{} by increasing the number of concurrent trajectory collections. In this experiment, we provision sufficient CPU resources (768 CPUs) and vary the number of H100 GPU nodes used by the system. \Cref{fig:env/pressure_test} (\textit{right}) shows the system exhibits strong scaling that is sub-linear, but very close to the desired linear speed-up curve. This slowdown primarily stems from the synchronization overhead from coordinating rollouts across many nodes.

\section{Training Web Agents with \ours{}} \label{sec:train}

We train web agents using the tasks and system provided by \ours{}. Our goal is to study the design choices behind a simple, visual web agent by training on \ours{}.  Concretely, we center our empirical results around two central questions.
First, we study how \ours{} can transform existing VLMs to efficiently solve complex web tasks via RL training (\S\ref{sec:train/enabling-long-horizon}).  
Second, we investigate how the agent’s generalization capability scales when trained with the large and diverse task set provided by \ours{}, which eventually leads to performance improvement (\S\ref{sec:train/scale-up}). We show that via a simple REINFORCE-style RL training procedure, we can train agents that improve performance from the initial 26.2\% to 42.9\% on the test split of \ours{}. The hyper-parameters of the best run are shown in~\S\ref{app:hparams}.

\subsection{Experimental Setup} \label{sec:train/setup}

To study the two questions outlined above, we first establish the experimental setup we operate in.

\textbf{Action space.}
We parameterize the action space by extending the definition of~\citet{test-time-interaction} with a navigation action, so that the action space now includes clicking, typing, scrolling, returning to the previous page, and navigating to a website specified by the agent. The \ours{} rollout system supports both the Set-of-Marks mode~\citep{set-of-marks}, where interactive elements are labeled in the screenshots by the simulator, and a coordinate-based screenshot-only mode. We operate directly in the coordinate-based mode, which is supported by the base model (Qwen3-VL-8B) we use to train via RL.

\textbf{Policy update.}
We run REINFORCE~\citep{original-reinforce} with binary terminal rewards, without any baseline or negative gradient. This update is equivalent to online filtered behavior cloning (BC)~\citep{digirl} or a ``thresholded'' version of reward-weighted regression~\citep{reward-weighted-regression}, which retains only successful trajectories and maximizes their log-likelihood. The training objective is given by:
\begin{align}
\argmax_{\theta}~~
\mathbb{E}_{\mathcal{T}\sim \text{tasks}}
\left[
\mathbb{E}_{o_{0:\tau},\, a_{0:\tau-1}\sim \pi_\theta(\cdot \mid \mathcal{T})}
\left[ 
\left(\sum_{t=0}^{\tau-1} \log \pi_{\theta}(a_{t} \mid o_{\leq t}, \mathcal{T})\right)
\cdot \mathds{1}\!\left[R(o_{0:\tau}, \mathcal{T}) =1 \right]
\right]
\right],
\end{align}
where $\theta$ denotes the parameters of the (multi-step) agent policy $\pi_\theta$, and $\mathcal{T}\sim\text{tasks}$ is a task instance sampled from the training task distribution. A rollout proceeds in discrete interaction steps indexed by $t$. At each step $t$, the agent samples an action $a_t \sim \pi_\theta(\cdot \mid o_{\le t}, \mathcal{T})$ conditioned on the full observation history $o_{\le t}$ (and the task), and the environment returns the next observation. The rollout terminates at step $\tau$ only when one of the following two conditions are met: \textbf{(i)} when the agent emits an \texttt{ANSWER} action, or \textbf{(ii)} when the step index reaches the training horizon $h_{\text{train}}$, the maximum allowed number of interaction steps during training; thus $\tau \le h_{\text{train}}$. The realized trajectory therefore contains actions $a_{0:\tau-1}$ and observations $o_{0:\tau}$, where $o_0$ is the initial observation and $o_\tau$ is the terminal observation after the last action. If termination occurs under the first condition (i.e., when an \texttt{ANSWER} action is produced), the environment does not advance and we set $o_\tau = o_{\tau-1}$ (i.e., the final observation is identical to the last pre-answer observation). The binary terminal reward $R(o_{0:\tau}, \mathcal{T})\in\{0,1\}$ indicates whether the overall trajectory succeeds on task $\mathcal{T}$. The objective increases the log-likelihood of all actions along successful trajectories while assigning zero weight to unsuccessful ones, avoiding negative gradients from failures. In principle, we expect that incorporating a baseline (negative gradient) would only improve performance further, but chose to avoid it for now for faster experimentation due to improved training stability with only a positive gradient. 

\textbf{Base policy.} We use the Qwen3-VL model family for its strong grounding and reasoning capabilities~\citep{qwen3-vl, qwen3vl-repo}. We experiment on the 8B-sized model with both Instruct and Thinking variants.

\textbf{Task sampling approach.} We ablate on random task sampling and ratio-over-difficulty sampling over the \ours{} task set, and present results for both approaches in~\S\ref{sec:train/scale-up}. We also maintain a dynamic blocklist of websites that prevent access, avoiding those sites for subsequent training periods (\S\ref{app:env/anti-blocking}).

\textbf{Controlling horizon during training.} To keep rollout collection efficient, we cap the maximum number of interaction steps per episode. Although these trajectories may still be \emph{successful} (i.e., they eventually solve the task), they can be \emph{suboptimal} in terms of step count, taking many unnecessary interactions compared to a shorter solution. Without a step cap, rollouts may include many such long-but-successful trajectories, which are expensive to collect and dilute training with avoidable steps. By truncating the horizon, we reduce wasted environment interactions during data collection and bias training toward shorter, more efficient successes, improving step efficiency. This is also preferable to simply tuning the learning objective to emphasize trajectories shorter than a threshold while still collecting long trajectories, because the horizon cap prevents spending those extra steps in the first place rather than only downweighting them after collection. We study the effect of the training horizon in~\S\ref{sec:train/scale-up}.

\textbf{Probing task hardness with proprietary VLMs.} To understand the hardness of the OOD test set in \ours{}, we also evaluate GPT-4o and GPT-5 (Thinking) as two strong proprietary models on the OOD test set. Results show that simply using the same prompt as Qwen3-VL (\S\ref{app:prompt/agent}) with Set-of-Marks (SoM~\citep{set-of-marks}, used here because these models cannot accurately output coordinates) leads to non-trivial but far below 100\% performance, where GPT-4o achieves 27.1\% on the full test set, and GPT-5 achieves 29.8\% on a randomly sampled subset with 300 tasks from the OOD test set (subsampled due to the budget costs associated with GPT-5). Note that GPT-4o achieves 25.6\% on this task subset, so the variance caused by subsampling is not significant. As we discuss next, our trained agent significantly outperforms both of these compared approaches.

\begin{figure*}[t!]
  \centering
  \setlength{\tabcolsep}{3pt}
  \begin{tabular}{@{}m{0.48\textwidth}m{0.48\textwidth}@{}}
    \begin{tikzpicture}
      \node[inner sep=0] (img) {\includegraphics[width=\linewidth]{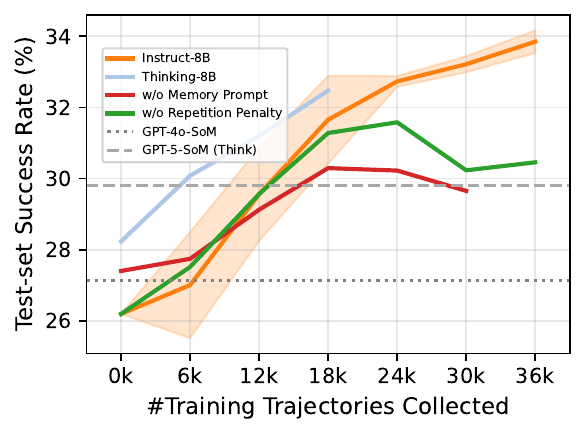}};
      \node[anchor=north west, fill=white, fill opacity=0.8, text opacity=1, inner sep=2pt] at (img.north west) {\textbf{(L)}};
    \end{tikzpicture} &
    \begin{tikzpicture}
      \node[inner sep=0] (img) {\includegraphics[width=\linewidth]{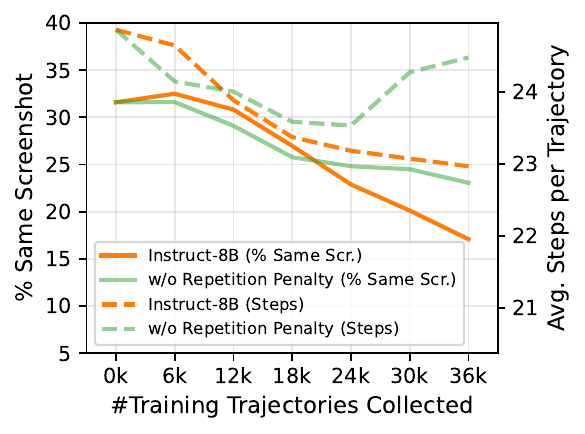}};
      \node[anchor=north west, fill=white, fill opacity=0.8, text opacity=1, inner sep=2pt] at (img.north west) {\textbf{(R)}};
    \end{tikzpicture}
  \end{tabular}
  \vspace{-0.25cm}
  \caption{\textbf{Ablations on base models, prompting, and action filtering.} \textit{(left)} test-set success rate curves of different models (Qwen3-VL-Instruct-8B, Qwen3-VL-Thinking-8B, GPT-4o, and GPT-5-Thinking), and of Qwen3-VL-Instruct-8B under different constraints (either removing the memory prompt or removing the repetition penalty during RL). \textit{(right)} test-set same-screenshot (repetitive inefficient action) rate and trajectory length curves before and after applying the action repetition penalty (filtering) on the Instruct model. The \textcolor{orange}{orange} curve is an average of two runs, while the translucent orange area around the orange line denotes the variance of these two runs. We observe that even if the starting variance between different runs can be large, the curves converges after 24k training trajectories are collected. All plots are produced with an EMA (Exponential Moving Average) smoothing of $0.50$.}
  \label{fig:ablations_ab}
  \vspace{-0.1cm}
\end{figure*}

\subsection{Design Choices for Agent Training} \label{sec:train/enabling-long-horizon}
In this subsection, we describe the key design choices we had to utilize to train agents via RL on \ours{}. Inspired by prior research on RL for LLMs~\citep{e3, reinforce-ada}, we note that biasing the sampled tasks towards \textit{hard tasks} improves the performance and sample efficiency, as long as easy tasks are also used either via co-training or via a prior stage of a curriculum. Thus, we start from a difficulty sampling that slightly biases towards harder tasks (concretely, easy : medium : hard = $2:5:3$). All experiments in this section follow this task sampling strategy, while we study different sampling strategies further in~\S\ref{sec:train/scale-up}.

\textbf{Memory prompt for long-horizon tasks.}
Our base model (Qwen3-VL-8B) follows a simple prompting template for computer-use agents, which includes a high-level action description (e.g., ``click on the search button'') paired with a low-level tool call (e.g., ``Click [209, 441]'')~\citep{qwen3-vl, qwen3vl-repo}. The exact prompt is shown in~\S\ref{app:prompt/agent}. From preliminary experiments, we observe that this template yields strong initial performance, because the Qwen-3 series of models are trained on data collected from web environments with this prompt~\citep{qwen3-vl, qwen3vl-repo}, but this prompt provides limited improvement if we use it to rollout trajectories with RL on \ours{}, as shown in~\Cref{fig:ablations_ab} (\textit{left}).

This limitation arises from the model’s inability to retain and reuse information gathered in earlier steps. For example, in the task shown in~\Cref{fig:env/decompose_task_example}, where the goal is to compare two products, the agent must recall details about the first product when evaluating the second. In the current design, this requires including the entire conversation history and all prior screenshots as input, which is inefficient and impractical. A more formal treatment of this challenge from a partially observable Markov decision process (POMDP) perspective is provided in \S\ref{app:modeling-pomdp}. To solve this limitation, we propose that the model should maintain a memory by responding with the updated memory in each step, inspired by GLM-4.1v~\citep{glm-4.1v}. The prompt with memory is also demonstrated in~\S\ref{app:prompt/agent}. Utilizing the memory prompt significantly boosts performance after RL (see \Cref{fig:ablations_ab}, \textit{left}, \textit{Instruct-8B} versus \textit{w/o Memory Prompt}).

\noindent\textbf{Removing repeated actions with a reward penalty.} We frequently observe cases where the base model is stuck at the same screenshot and attempts to repeat the same action multiple times. In order to measure how often this behavior happens, we calculate the fraction of times the screenshot at one step is the same as the next one (\textit{\% Same Screenshot}), and the trajectory length (\textit{Avg. Steps per Trajectory}) in~\Cref{fig:ablations_ab} (\textit{middle}). Preliminary experiments show that standard RL substantially reduces this behavior of repeating ineffective actions, but does not fully eliminate it(\Cref{fig:ablations_ab}, \textit{middle}). Thus, we design a penalty for these repeated actions when doing RL: we explicitly filter out a step if the resulting next step presents an identical screenshot, even if the current step is a part of a successful trajectory. After applying this penalty, our RL runs become substantially more sample efficient as shown in \Cref{fig:ablations_ab}, \textit{left} (Instruct 8B vs ``w/o Repetition Penalty'').

\noindent\textbf{Use of thinking for every step.} In addition to the \emph{Instruct} variant of Qwen3-VL-8B, we also experiment with the \textit{Thinking} variant of this model with exactly the same prompt as the Instruct model. Results as shown in~\Cref{fig:ablations_ab} (\textit{left}) reveal that the Thinking model attains a higher initial performance compared to the Instruct model, suggesting that extended reasoning at each step can be beneficial. However, the Thinking model produces significantly longer responses: before RL training: for instance, it produces \textbf{2139} characters per response compared to \textbf{1088} of the Instruct model on the held-out test tasks. Thus training the Thinking model incurs a larger latency cost and requires more compute. Notably, as training progresses with \ours{}, the performance gap between the Instruct and Thinking models decreases considerably, and soon the Instruct model surpasses the Thinking model. Given the favorable trade-off between performance and efficiency, we adopt the Instruct model for subsequent experiments, though we believe that developing a compute-efficient RL recipe to effectively train a Thinking model is an interesting direction for future work.

\subsection{Main Results: Scaling Agent Performance with \ours{}} \label{sec:train/scale-up}

\begin{table}[t!]
  \centering
  \setlength{\tabcolsep}{6pt}
  \renewcommand{\arraystretch}{1.15}
  \begin{tabular}{l l c c c}
    \toprule
    & \textbf{Base Model} & \textbf{Prompt} & \textbf{RL on WebGym} & \textbf{Performance (\%)} \\
    \midrule
    \multirow{2}{*}{\textbf{Proprietary Models}}
      & \texttt{GPT-4o} & Memory & \textcolor{gray}{\xmark} & \gcmaintable{27.1} \\
      & \texttt{GPT-5}  & Memory & \textcolor{gray}{\xmark} & \gcmaintable{29.8} \\
    \midrule
    \multirow{4}{*}{\textbf{Open-source Models}} 
      & \texttt{Qwen3-VL-Instruct-8B} & Official* & \textcolor{gray}{\xmark} & \gcmaintable{27.4} \\
      & \texttt{Qwen3-VL-Instruct-8B} & Memory   & \textcolor{gray}{\xmark} & \gcmaintable{26.2} \\
      & \texttt{Qwen3-VL-Thinking-8B} & Memory   & \textcolor{gray}{\xmark} & \gcmaintable{28.2} \\
      & \texttt{Qwen3-VL-Instruct-8B} & Memory   & \textbf{\cmark (Ours)} & \cellcolor{high!100!low!\opacity}$\mathbf{42.9}$ \\
    \bottomrule
  \end{tabular}
  \caption{\textbf{\ours{} empowers the small \texttt{Qwen3-VL-Instruct-8B} model to learn with RL to achieve state-of-the-art performance on the holdout test task set}, arriving at a strikingly 42.9\% success rate, significantly surpassing all agents empowered by proprietary models and small-yet-effective open-source agentic models. *The \textit{Official} prompt refers to the web agent prompt specified in~\citet{qwen3vl-repo}.}
  \label{tab:main_table}
\end{table}

Next, we present our main results scaling agent performance with \ours{}. In particular, we are interested in understanding \emph{what} concretely about the design of \ours{} scales performance the most, and how much improvement we can obtain by running RL on \ours{}. In the end, we're able to scale the performance up to 42.9\%, significantly surpassing existing agents empowered by proprietary models like GPT-4o and GPT-5-Thinking and strong open-source models like Qwen3-VL-Instruct-8B and Qwen3-VL-Thinking-8B. The comparison of these results are demonstrated in~\Cref{tab:main_table}. In this section, we will understand how \ours{} enables these results by specifically studying the effects of: \textbf{(i) }\textit{breadth} (domain diversity), \textbf{(ii)} \textit{depth} (difficulty composition), and \textbf{(iii)} the train-time \textit{interaction horizon}, during RL training on \ours{}. Our starting point is the best checkpoint from~\S\ref{sec:train/enabling-long-horizon}, which \textit{biases towards harder tasks} (easy : medium : hard = $2:5:3$).

\begin{figure*}[t!]
  \centering
  \includegraphics[width=0.85\linewidth]{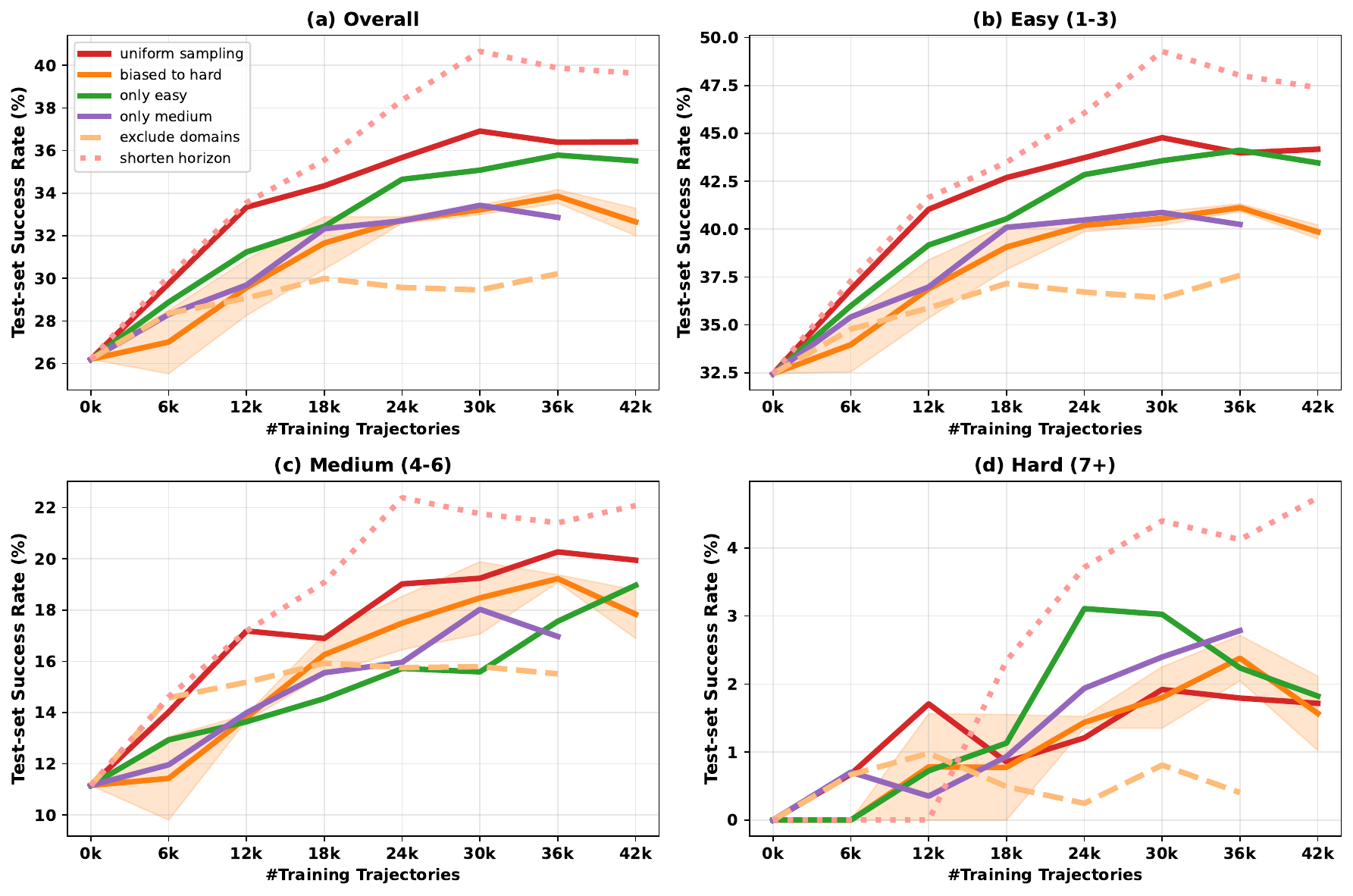}
  \caption{\textbf{Exploring scaling dimensions of \ours{} and difficulty-aware training.} Test-set success rate curves under different variations to training: removing domains (``exclude domains''), tuning difficulty ratios (``uniform sampling'', ``biased to hard'', ``only easy'', ``only medium''), and shortening the train-time step budget (``shorten horizon''). Results are reported for \textbf{(a) Overall} (all tasks), \textbf{(b) Easy (1--3)}, \textbf{(c) Medium (4--6)}, and \textbf{(d) Hard (7+)}. The \textcolor{orange}{orange} curve averages two runs; the translucent orange band shows their variation. All plots are produced with an EMA (Exponential Moving Average) smoothing of $0.50$.}
  \label{fig:ablations_cde}
\end{figure*}

\textbf{1) Scaling task set \textit{breadth} (domain coverage) for better generalization.}
To study the effect of having a wide variety of domains in \ours{}, we randomly remove half of the subdomains from the training set (``exclude domains'') while keeping the same training recipe otherwise as all the comparisons. This results in the subsampled training task set that only contains 53\% of original tasks. Across all evaluation slices, removing domains (``exclude domains'') compared to using all tasks from \ours{} (``biased to hard'') consistently slows down improvements and lowers the final success rate (\Cref{fig:ablations_cde}a--d), indicating that domain breadth matters even when the difficulty mixing ratio is unchanged.
This gap is not due to oversampling and overfitting on fewer tasks: note that the number of collected training trajectories (36k) is far smaller than the number of training tasks remaining ($\sim$ 150k), so most training happens on distinct tasks. The degradation reflects reduced domain diversity, which limits generalization to unseen websites and subdomains.

To study 2) and 3) below, we compare four task compositions: \textbf{a)} training only on easy tasks (``only easy'', $1:0:0$), \textbf{b)} only on medium tasks (``only medium'', $0:1:0$), \textbf{c)} a task mixture biased towards harder tasks (``biased to hard'', $2:5:3$, as used in \S\ref{sec:train/enabling-long-horizon}), and \textbf{d)} sampling tasks uniformly at random from the full training set (``uniform sampling'', approximating a ratio of $25:5:1$ given the task set).

\textbf{2) Scaling candidate task set \textit{size} by sampling more easy tasks avoids overfitting to narrow hard tasks with the filtered BC loss.} In our previous experiments, we sample with a difficulty mixture of $2:5:3$, which we note is heavily biased towards harder tasks compared to uniform sampling, because easy tasks actually comprise around $80\%$ of all tasks. This can make the effective task set size (i.e., expected number of tasks sampled at any given training iteration that are distinct from all current and previous iterations) very small because a large portion of the training tasks are repeated high-difficulty tasks. As we roll out more and more trajectories, we believe that this imbalance leads to overfitting because the test performance curve begins to drop right around 36k-42k training trajectories on the x-axis, as shown in~\Cref{fig:ablations_cde} (\textit{biased to hard}). This problem can, in principle, be resolved with a better loss function or, perhaps more simply, be resolved by scaling the size of candidate task set. We choose to scale up the size of  the candidate task set by adjusting the task sampling ratio so that the sampled tasks include more easy tasks that are distinct from each other.

In fact, we found that training exclusively on easy tasks performs surprisingly well and is more stable than over-emphasizing medium and hard tasks. In particular, the ``only easy'' setting consistently outperforms “only medium” on overall success as well as on easy and medium test tasks (\Cref{fig:ablations_cde}a–c), and it does not exhibit the late-stage plateauing or mild performance regression observed when harder tasks are upweighted (e.g., the “biased to hard” setting after $\sim$36k trajectories; \Cref{fig:ablations_cde}a–c).
This behavior contrasts with typical findings in LLM reasoning, but is consistent with the structure of \ours{}, where the very large set of easy tasks spans many websites and domains and thus supports generalization to unseen websites. On the other hand, very hard tasks in math reasoning can be built for all sub-areas for mathematics. While repeated exposure to a narrower set of domains that provide harder tasks remains important (as discussed below), the size afforded by easy tasks plays a crucial role in mitigating overfitting and enabling generalization to new websites. In contrast, aggressively upweighting medium and hard tasks increases repeated exposure to a smaller subset of domains and interaction patterns, leading to overfitting that manifests as performance plateaus and slight degradation in our results.

\textbf{3) Scaling candidate task set \textit{depth} (wider range of difficulty levels) by including tasks from all difficulty levels to promote more efficient learning.} Despite the strong performance of the “only easy” setting, we believe that higher-difficulty tasks still play an important role. This intuition is also supported by our empirical observations: the performance gap between “only easy” and “biased to hard” is substantially smaller on the medium- and hard-difficulty test subsets than on the easy-difficulty subset.
Motivated by this, we increase the depth of the task set by incorporating tasks spanning a wider range of difficulty levels. The simplest way to achieve this is through \textit{uniform sampling} from the full training set. We find that uniform sampling yields the highest overall performance among the four difficulty-mixing strategies we evaluate (\Cref{fig:ablations_cde}a), and it achieves a significantly higher success rate on the \textit{medium-difficulty} test set compared to training on easy tasks alone (\Cref{fig:ablations_cde}c). Because the “only easy” setting already draws from a large and diverse pool of tasks, the additional gains from uniform sampling are best attributed to the inclusion of some medium- and hard-difficulty tasks. Injecting these harder tasks provides training signals that enable the agent to develop capabilities beyond easy interactions, particularly on medium-difficulty test tasks. Finally, since uniform sampling includes only a small fraction of hard tasks, we do not observe a significant difference between these two settings on the hard-difficulty test set.

\textbf{4) Horizon: controlling train-time interaction budget.}
Additionally, we study one important train-time hyperparameter for algorithmic exploration: the train-time interaction budget. To control this budget and shape the distribution of training trajectories, we reduce the per-episode step limit from $(15, 30, 45)$ to $(10, 20, 30)$ (``shortened horizon''). Empirically, shortening the horizon improves both sample efficiency and final performance across all difficulty slices. In particular, the success rate increases from a peak of 38.2\% at the previous best checkpoint (uniform sampling with a step budget of $(15, 30, 45)$) to a peak of 42.9\%.
We hypothesize that a tighter step budget acts as a regularizer on the policy’s interaction process. By limiting opportunities for late-stage recovery, it discourages trajectories that succeed only after extended sequences of low-yield exploration, and instead concentrates training signal on earlier, higher-impact decisions. This increases the density of informative state–action transitions per episode, improving credit assignment and reducing variance in policy updates. Notably, performance on hard tasks also improves substantially (\Cref{fig:ablations_cde}d), consistent with the model learning more robust and reusable interaction primitives (e.g., efficient navigation and decisive goal completion) that transfer beyond the capped training horizon and support longer-horizon problem instances at evaluation time.

\begin{table}[t!]
\centering
\begin{tabular}{llcc}
\toprule
\textbf{Scaling} & \textbf{Legend in~\Cref{fig:ablations_cde}} & \textbf{Peak Performance (\%)} & \textbf{$\Delta$ from Previous (\%)} \\
\midrule
Baseline & Exclude Domains & 31.0 & -- \\
+Breadth & Biased to Hard & 34.5 & +3.5 \\
+Size & Only Easy & 36.9 & +2.4 \\
+Depth & Uniform Sampling & 38.2 & +1.3 \\
+Horizon Control & Shorten Horizon & 42.9 & +4.8 \\
\bottomrule
\end{tabular}
\caption{\textbf{Scaling overview: peak performance of different scaling directions through RL settings on the test set. }$\Delta$ from Previous shows the improvement over the immediately preceding setting.}
\label{tab:scaling-overview}
\end{table}

All main results from this section are summarized in~\Cref{tab:raw-results}, and the overview of the scaling results we discussed above is presented in~\Cref{tab:scaling-overview}. Our final agent is trained with the memory prompt and penalty for repetitive actions, with a task horizon of $(10,20,30)$ over (easy, medium, hard) tasks, and with uniform sampling across all task difficulties, surpassing the best proprietary model evaluated (GPT-5-Thinking) by 13.1\%, as shown in~\Cref{tab:main_table} and~\Cref{tab:raw-results}. We provide raw results to promote fair comparison, and encourage readers to include raw results in future work to compare with these scores.

\section{Discussion and Conclusion}

We introduced \textit{WebGym}, the largest open-source training environment for visual web agents. Our procedural task construction approach generates structured evaluation rubrics and decomposes seed tasks into valid criterion subsets, enabling both domain breadth and hardness depth while ensuring decomposed tasks remain semantically coherent. To support large-scale training, we developed an asynchronous rollout system that eliminates synchronization barriers, substantially accelerating data collection for on-policy RL.

Our experiments reveal several empirical insights: explicit memory mechanisms are essential for long-horizon tasks requiring cross-step information retention; penalizing repeated ineffective actions improves sample efficiency; domain diversity directly translates to out-of-distribution generalization; uniform difficulty sampling outperforms both easy-only and hard-biased curricula; and shorter training horizons improve both efficiency and final performance. With these design choices, a simple REINFORCE algorithm enables strong generalization to entirely unseen websites, outperforming proprietary models.

\textbf{Future work.} Despite the availability of meaningful task-specific rubrics in our task set, LLM-generated rubrics can sometimes be overly strict, which slightly reduces sample efficiency during training. An important direction for future work is to explore more advanced designs for rubric-based evaluation. One promising approach is to treat LLM-generated rubrics as soft guidance rather than strict criteria, for example by training an evaluator that jointly conditions on both the rubric and the task instruction. Another promising direction is to investigate reinforcement learning algorithms that utilize \ours{} more efficiently, particularly methods based on dynamic curricula~\citep{test-time-interaction}, automatic curriculum learning~\citep{digirl}, multi-agent training, and cross-site navigation.

\section*{Acknowledgements}
We thank Rui Yang, Nan Jiang, Rui Pan, Yifei He from UIUC, Yifei Zhou from UC Berkeley, Anikait Singh, Salman Abdullah from Stanford for their early discussions of this research, members of the CMU AIRe lab for their feedback on human evaluation and presentations, and Eduardo Salinas, Gustavo De Rosa, Ahmed Awadallah from Microsoft Research (AI Frontiers Lab) for their continued support throughout this work. This work is partially supported by NSF under Grant No.\ 2416897, Grant No.\ 2505932, and by ONR under Grant No. N000142512318. This research used both Delta (NSF award OAC 2005572) and DeltaAI (NSF award OAC 2320345) advanced computing systems, and computing resources provided by Illinois Computes and NAIRR Pilot NAIRR250157.

{
    \small
    \bibliographystyle{plainnat}  % Changed from ieeenat_fullname to plainnat for compatibility
    \bibliography{main}
}

\clearpage
\appendix

% Note: AI Frontiers uses single column format, so no need for \onecolumn/\twocolumn
\section{Raw Experimental Results}
\label{app:raw-results}

We demonstrate the raw values of all experiments presented in~\S\ref{sec:train} in~\Cref{tab:raw-results} below.

\begin{table}[htbp]
  \centering
  \resizebox{\textwidth}{!}{%
  \begin{tabular}{lccccccc}
  \toprule
  \textbf{Method} & \textbf{$H_\text{train}$} & \textbf{Sampling} & \textbf{Domain} & \textbf{Memory} & \textbf{Rep. Penalty} & \textbf{Zero-shot (\%)} & \textbf{RL Peak (\%)} \\
  \midrule
  Instruct-8B & $15:30:45$ & $2:5:3$ & Complete & \cmark & \cmark & \gcperf{26.2} & \gcperf{34.5}$\pm 0.4$ \\
  Thinking-8B & \textcolor{gray}{$15:30:45$} & \textcolor{gray}{$2:5:3$} & \textcolor{gray}{Complete} & \textcolor{gray}{\cmark} & \textcolor{gray}{\cmark} & \gcperf{28.2} & -- \\
  \midrule
  W/o Memory Prompt & \textcolor{gray}{$15:30:45$} & \textcolor{gray}{$2:5:3$} & \textcolor{gray}{Complete} & \xmark & \textcolor{gray}{\cmark} & \gcperf{27.4} & \gcperf{31.5} \\
  W/o Repetition Penalty & \textcolor{gray}{$15:30:45$} & \textcolor{gray}{$2:5:3$} & \textcolor{gray}{Complete} & \textcolor{gray}{\cmark} & \xmark & \gcperf{26.2} & \gcperf{33.0} \\
  \midrule
  Uniform Sampling & \textcolor{gray}{$15:30:45$} & $25:5:1$ & \textcolor{gray}{Complete} & \textcolor{gray}{\cmark} & \textcolor{gray}{\cmark} & \gcperf{26.2} & \gcperf{38.2} \\
  Only Easy & \textcolor{gray}{$15:30:45$} & $1:0:0$ & \textcolor{gray}{Complete} & \textcolor{gray}{\cmark} & \textcolor{gray}{\cmark} & \gcperf{26.2} & \gcperf{36.9} \\
  Only Medium & \textcolor{gray}{$15:30:45$} & $0:1:0$ & \textcolor{gray}{Complete} & \textcolor{gray}{\cmark} & \textcolor{gray}{\cmark} & \gcperf{26.2} & \gcperf{35.0} \\
  \midrule
  Exclude Domains & \textcolor{gray}{$15:30:45$} & \textcolor{gray}{$2:5:3$} & Half & \textcolor{gray}{\cmark} & \textcolor{gray}{\cmark} & \gcperf{26.2} & \gcperf{31.0} \\
  Shorten Horizon & $10:20:30$ & $25:5:1$ & \textcolor{gray}{Complete} & \textcolor{gray}{\cmark} & \textcolor{gray}{\cmark} & \gcperf{26.2} & \cellcolor{high!100!low!\opacity}$\mathbf{42.9}$ \\
  \midrule
  \rowcolor{gray!20} GPT-4o-SoM & -- & -- & -- & \cmark & -- & \gcperf{27.1} & -- \\
  \rowcolor{gray!20} GPT-5-SoM (Think) & -- & -- & -- & \cmark & -- & \gcperf{29.8} & -- \\
  \bottomrule
  \end{tabular}%
  }
    \caption{\textbf{Raw ablations and performance results of our trained agents on the OOD test set.} $H_\text{train}$: training horizon (easy:medium:hard steps); Sampling: difficulty ratio (easy:medium:hard); Domain: subdomain coverage; Memory: memory prompt; Rep.\ Penalty: repetition penalty. For Instruct-8B (Biased to Hard), we report the peak value of the averaged curve from two runs, with $\pm$ indicating the half-range.}
    \label{tab:raw-results}
\end{table}

\section{Training Environment Designs}
\label{app:env}

\subsection{Rollout System: Operation-specific Local Request Queue} \label{app:env/local-queue}

To avoid flooding the server with requests, the most naive approach is to implement a global queue (first-in-first-out rule) to avoid flooding the server with requests. However, below we show the ineffectiveness of this implementation because (1) it causes extensive warmup time and (2) it creates operation bottlenecks. To improve it, we demonstrate that each operation type should have its own queue. This is to say, requests with different types of operations should not have any priority over each other.

\begin{figure*}[htp!]
    \centering
    \includegraphics[width=\linewidth]{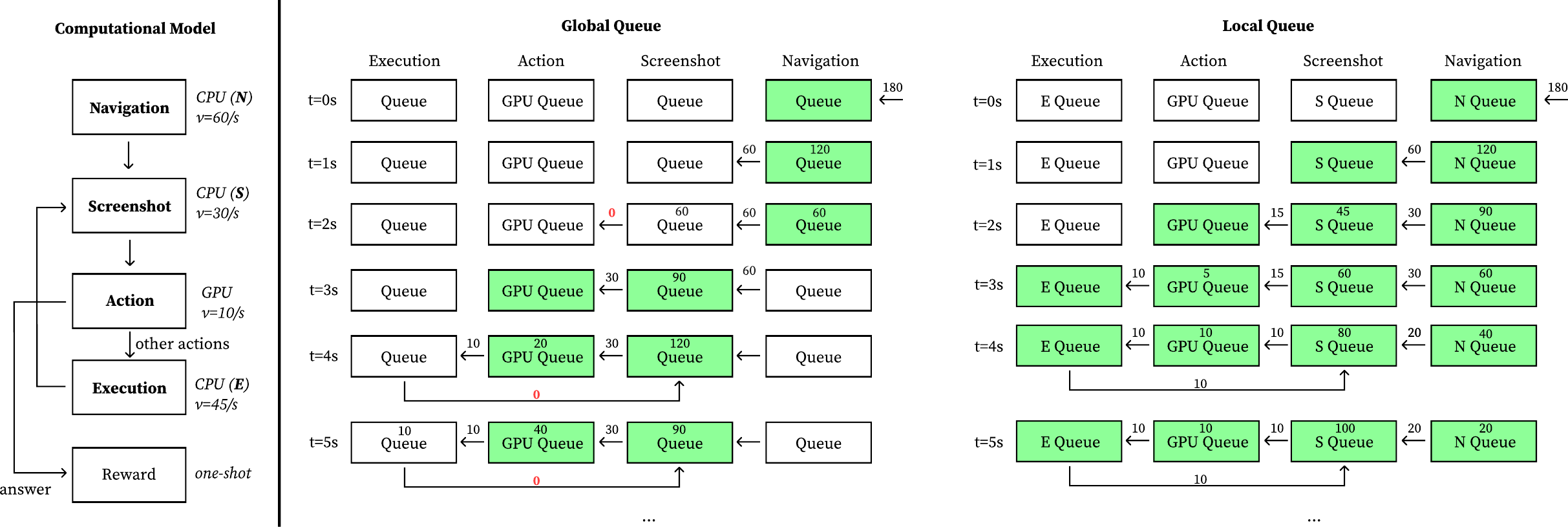}
    \caption{\ours{} implements an \textbf{\textit{operation-specific queue system}} that balances the CPU and GPU machine usages. Here we illustrate the computational model of this pool system in the \textit{left} subplot, and the comparison of the two queue designs on the \textit{right}. We represent the centralized CPU queue with \textit{Queue}, and the cascading queue system for with the initial letter of the request type, specifically, Navigating to a webpage (\textbf{N}), taking a Screenshot (\textbf{S}), and Executing Action (\textbf{E}) that takes medium amount of time. A green box means the CPU machine is under optimal load.}
    \label{fig:env/local_queue}
\end{figure*}

The computational model of rolling out a trajectory in a multi-step RL setting is shown in~\Cref{fig:env/local_queue} (\textit{Left}). When simulation is required for an RL environment, there will be a sequence of CPU operations bounded by CPU resources. On the other hand, high-speed RL inference usually put high pressure on GPU resources as well. For the web agents rollout collection, the computational model can be simplified to five stages: navigating to a website with the simulator (N), taking a screenshot with the simulator (S), proposing an action with the agent, executing an action with the simulator, and sending the trajectory to an evaluator to obtain reward (E), with their speed defined on the right side of each box.

Assume we send a burst batch of 180 requests to the rollout system at $t=0s$. As shown in~\Cref{fig:env/local_queue} (middle), with a global queue, at $t=1s$, although 60 Navigation requests are finished, the Screenshot requests arrive \textit{strictly after} the rest of the Navigation requests, so they have to wait for all Navigation requests to return. This makes GPU \textbf{starve} at $t=2s$. Similarly, after $t=5s$, only after all Screenshot requests are returned can the Execution requests be processed, and after all Execution requests can be processed can the next-step Screenshot requests be processed, causing the GPU to \textbf{starve again and again} during the multi-step loop.

To improve this, \ours{} implements a local queue that does not specify any priority \textit{among} operations. At $t=1s$, Screenshot requests and Navigation requests will be processed with the same priority. This results in 30 Navigation requests and 15 Screenshot requests being returned at $t=2s$ (as they share CPU resources). When propagating this pattern through the pipeline, the GPU resources will not starve at any moment.

\begin{figure}[t!]
  \centering
  \begin{subfigure}[c]{0.48\linewidth}
    \centering\vspace{0pt}
    \includegraphics[width=\linewidth]{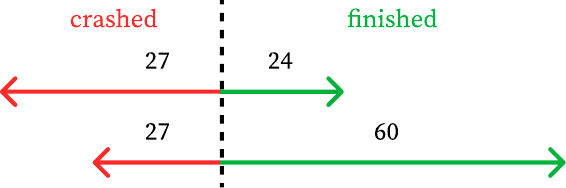}
  \end{subfigure}\hfill
  \begin{subfigure}[c]{0.50\linewidth}
    \centering\vspace{0pt}
    \includegraphics[width=\linewidth]{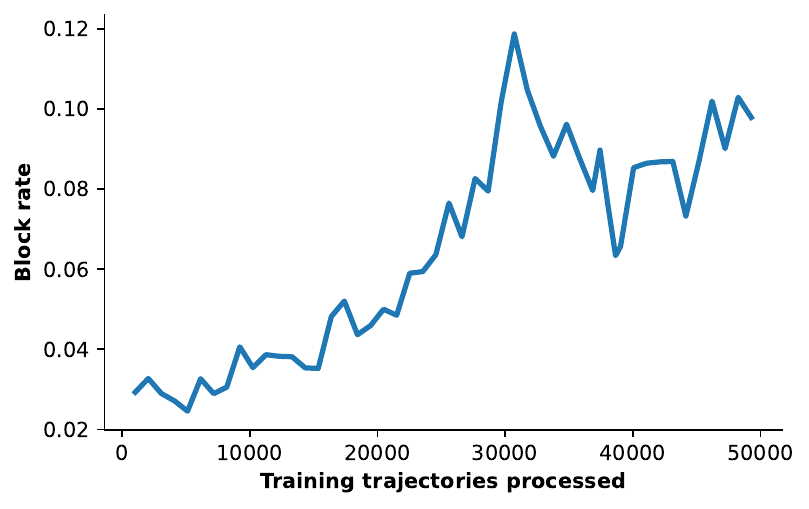}
  \end{subfigure}
  \caption{Left: Number of trajectories successfully finished and crashed out under high CPU load when $t=10 \text{min}$ with 64 CPUs, 256 environments, and 3 GPU nodes. Top line: \textit{global} queue. Bottom line: \textit{local} queue. Right: \textbf{The block rate increases as training goes on.}}
  \label{fig:app/env}
\end{figure}

We experiment with the performance boost of this local queue under extreme CPU condition (64 CPUs needs to run 256 environments). Results are shown in~\Cref{fig:app/env} (left). We observe that the crash percent decreases significantly if we employ the local queue, showcasing the effectiveness of this design.

\subsection{Rollout Sampler: Anti-blocking} \label{app:env/anti-blocking}

As we observe that some websites block frequent requests from the same server as shown in~\Cref{fig:app/env} (right), we maintain a list of websites that blocks the agent with an LLM. The blocking is detected by GPT-4o with the prompt specified in~\S\ref{app:prompt/eval/block}.

\section{Modeling the Markov Decision Process On Long-Horizon Web Navigation Tasks}
\label{app:modeling-pomdp}

A partial observable Markov decision process (POMDP) extends a Markov decision process (MDP) to settings where the agent does not have direct access to the underlying state but instead receives incomplete observations. Formally, a POMDP is defined as a tuple \( (S, A, P, R, O, \Omega, \gamma) \), where \( S \) is the set of latent states, \( A \) is the action space, \( P(s' \mid s, a) \) defines the state transition dynamics, \( R(s, a) \) is the reward function, \( O \) is the observation space, \( \Omega(o \mid s) \) specifies the observation model that maps hidden states to observations, and \( \gamma \in [0,1) \) is the discount factor. Because the agent does not directly observe \( s_t \), it maintains a belief state \( b_t \), a probability distribution over \( S \), which it updates using past actions and observations. As a result, the optimal policy maps belief states to actions, \( \pi(b_t) \rightarrow a_t \), instead of mapping fully observed states to actions as in standard MDPs.

Recent work models the web agent tasks as a POMDP, because historical observations must be included to successfully infer the next action. Formally, for a web agent task, if we represent the observation (including the screenshot and metadata of that webpage) at step $t$ as $o_t$, recent work represents the true state $s_t$ with $s_t=o_{t-n:t}$ (concatenation of the observations of last $n$ steps)~\citep{qwen3-vl, test-time-interaction}. This is called a \textbf{windowed history}, designed to fit into the context limit of the vision-language models. Formally, this modeling assumes the approximation

\begin{equation} \label{equ:fixed-history-window-pomdp}
    \Pr\bigl( o_{t+1} | o_{\mathbf{1}:t}, a_{\mathbf{1}:t-1}, c \bigr) \approx \Pr\bigl( o_{t+1} | o_{\mathbf{t-n}:t}, a_{\mathbf{t-n}:t-1}, c \bigr),
\end{equation}

where $o$ is the vector that represents observation, and $c$ is the vector that represents task~\citep{pae-webvoyager, test-time-interaction, qwen3-vl, qwen3vl-repo}. In words, these works treat observations of the last several steps as sufficient statistics of history for predicting future observations. In practice, the window size $n$ is usually set to $3$. For long-horizon tasks, this modeling over-simplifies the problem because the policy will frequently return to the same state. If the policy is not aware of its full history, it can produce repetitive sequences of actions because it is not aware of what it already finished earlier.

Previously, there have been clever designs in token-level RL (modeling a token as a step), e.g. RL from human feedback (RLHF), that mitigate this problem by directly represent $s_t$ with the representation of the token from the auto-regressive vision-language model, which automatically encodes information from all previous steps, naturally enforcing $s_t=o_{1:t}$. Is there a way we can bring this into sequence-level RL (modeling a sequence as a step)?

Practically, the problem with inputting full history observations $o_{1:t}$ to the VLM policy is that the context window of even the most advanced VLMs is not large enough, partially because image is too dense a representation for web agents tasks. Recent work like GLM-4.1v~\citep{glm-4.1v} addresses this limitation by introducing a \textbf{memory} as part of the tokens that the VLM should output in each step. Formally, the memory at step $t$, denoted by $m_{t}$, is a result of the observation without memory $\tilde{o}_t$, the memory from the previous step $m_{t-1}$, and the task instruction $c$, denoted as $m_t=\text{VLM}_\theta(\tilde{o}_t, m_{t-1}, c)$, and the observation at step $t$ can now be denoted as $o_t=(m_t, \tilde{o}_t, c)$. This design leads to a continuous compression of the memory based on the progress of the task, which resembles RLHF and Long-Short-Term Memory (LSTM). We practice this idea for our agent context management, as shown in~\Cref{app:prompt/agent}.

% \section{Task Set Licenses}
% \label{app:task-set-details}

% % ---- Separate license table ----
% \begin{table}[htp!]
% \centering
% \caption{Licenses for included task sets.}
% \vspace{1mm}
% \label{tab:swe-gym-licenses}
% \begin{tabular}{ll}
% \toprule
% Dataset (split) & License \\
% \midrule
% Mind2Web-Live \citep{mind2web-live} & CC-BY-4.0 \\
% WebVoyager \citep{webvoyager} & Apache 2.0 \\
% Online Mind2Web \citep{online-mind2web} & CC-BY-4.0 \\
% InSTA-v3 \citep{insta-v3} & MIT \\
% AgentSynth-Web \citep{agentsynth} & Apache 2.0 \\
% BrowseComp \citep{browsecomp} & MIT \\
% GAIA-Web \citep{gaia} & Apache 2.0 \\
% Mind2Web-2 \citep{mind2web-2} & CC-BY-4.0 \\
% TravelPlanner \citep{travelplanner} & CC-BY-4.0 \\
% DeepShop \citep{deepshop} & CC-BY-4.0 \\
% PAE-WebVoyager \citep{pae-webvoyager} & CC-BY-4.0 \\
% \bottomrule
% \end{tabular}
% \end{table}

\section{Qualitative Examples} \label{app:qual}

\subsection{Example Rubrics} \label{app:qual/rubric}

We show some example rubrics generated in~\Cref{tab:qual/rubric}. We can observe a general correlation between the instruction length of the task and the number of criteria in its rubrics.

\setlength{\tabcolsep}{3pt}
\begin{table}[H]
\footnotesize
\centering
\caption{Examples of WebGym Tasks with Evaluation Rubrics}
\label{tab:qual/rubric}
\begin{tabular}{|p{0.32\textwidth}|p{0.67\textwidth}|}
\hline
\textbf{Task} & \textbf{Evaluation Rubric} \tabularnewline
\hline
Find out the date of the next Baden Marathon event. & \textbf{Group 1:} Find date of next Baden Marathon event \newline \textit{Facts:} \newline - date of the next Baden Marathon event \tabularnewline
\hline
Find the recommended daily feeding amount for an adult British Shorthair cat (over 12 months) on royalcanin.pl, assuming a target weight of 5kg and a moderate activity level. & \textbf{Group 1:} Identify the target cat specifications \newline \textit{Facts:} \newline - adult British Shorthair cat \newline - age over 12 months \newline - target weight of 5kg \newline - moderate activity level \newline \textbf{Group 2:} Find recommended daily feeding amount \newline \textit{Facts:} \newline - recommended daily feeding amount for the specified cat on royalcanin.pl \tabularnewline
\hline
Find the highest-rated mover in Honolulu to shift a vehicle and large appliances and who has virtual discussion options available in Yelp. & \textbf{Group 1:} Identify highest-rated mover in Honolulu \newline \textit{Facts:} \newline - mover is located in Honolulu \newline - mover is highest-rated \newline \textbf{Group 2:} Services provided by the mover \newline \textit{Facts:} \newline - mover offers vehicle shifting services \newline - mover offers large appliances shifting services \newline \textbf{Group 3:} Communication options available \newline \textit{Facts:} \newline - mover has virtual discussion options available \newline \textbf{Group 4:} Source of information \newline \textit{Facts:} \newline - information is found on Yelp \tabularnewline
\hline
\end{tabular}
\end{table}

\subsection{Comparing Rubric-based Evaluation with Task-based Evaluation} \label{app:qual/eval}

\begin{figure}[H]  % r or l for right or left
  \vspace{-0.5\baselineskip}          % pull it up a bit (optional)
  \centering
  \includegraphics[width=0.60\textwidth]{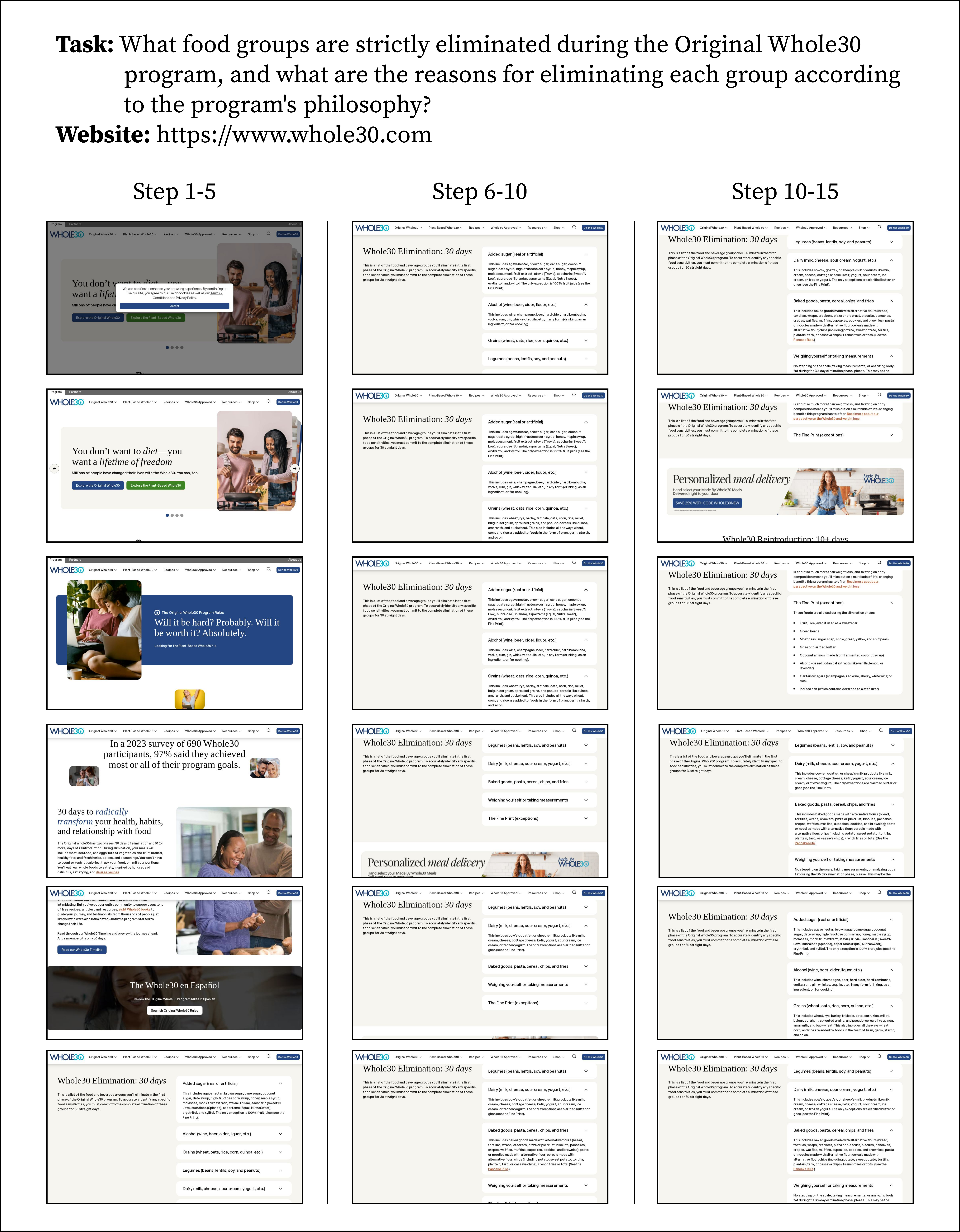}
  \caption{Trajectory for the comparison between vanilla evaluation and rubric-based evaluation.}
  \label{fig:qual/eval}
\end{figure}

We find many cases where rubric-based evaluation is able to produce significantly more comprehensive evaluation that checks details of each criteria of the task, which leads to many wins over using only the task description. One example is shown in~\Cref{fig:qual/eval}. The human evaluation for this example is \textcolor{successgreen}{NOT SUCCESS}.

\noindent \textbf{Task description-based evaluation.} 

\begin{itemize}
    \item \textbf{Verify criterion a:} The response mentions the food groups eliminated during the Original Whole30 program: added sugar, alcohol, grains, legumes, dairy, and baked goods, pasta, cereal, chips, and fries. The screenshots provide evidence of these eliminations as listed on the Whole30 website. Additionally, the reasons mentioned in the response for eliminating these groups, such as identifying food sensitivities and improving relationship with food, are also corroborated by the screenshots, particularly those discussing the philosophy of Whole30.
    \item \textbf{Verify criterion b:} The task asked for the food groups strictly eliminated during the Original Whole30 program and the reasons for their elimination according to the program's philosophy. The agent's response matches the task requirements accurately, listing the correct food groups and providing the reasons as asked, based on the information available in the screenshots.
    \item \textcolor{red}{\textbf{Verdict: SUCCESS}}
\end{itemize}

\noindent \textbf{Rubric-based evaluation.}

\begin{itemize}
    \item There are 2 groups for this rubric in total.

    \item \textbf{Group 1: Identify food groups eliminated during the Original Whole30 program}
    \item \begin{itemize}
        \item \textbf{Description:} Identify food groups eliminated during the Original Whole30 program
        \item \textbf{Facts:} Identify the food groups strictly eliminated during the Original Whole30 program
        \item \textbf{Verify criterion a:} The response claims that the food groups strictly eliminated during the Original Whole30 program include added sugar, alcohol, grains, legumes, dairy, and items such as baked goods, pasta, cereal, chips, and fries. The screenshots clearly show a list of eliminated food groups that includes added sugar (real or artificial) alongside the other listed categories. Therefore, the response’s identification of eliminated food groups is supported by the screenshots.
        \item \textbf{Verify criterion b:} This group requires identifying the food groups strictly eliminated during the Original Whole30 program. The response provides a concrete list of eliminated groups (including added sugar), which directly satisfies the identification requirement.
        \item \textbf{Verdict:} SUCCESS
    \end{itemize}

    \item \textbf{Group 2: Reasons for eliminating each food group}
    \item \begin{itemize}
        \item \textbf{Description:} Reasons for eliminating each food group
        \item \textbf{Facts:} Reasons for eliminating each food group according to the Original Whole30 program's philosophy
        \item \textbf{Verify criterion a:} While the screenshots list the eliminated food groups, they do not explicitly provide the program’s philosophy-based reasons for eliminating each group. The response offers general motivations (e.g., identifying sensitivities or resetting habits), but these statements are not directly verifiable from the screenshots as provided.
        \item \textbf{Verify criterion b:} This group specifically asks for reasons for eliminating \emph{each} food group according to the Original Whole30 philosophy. The response does not provide distinct, food-group-specific reasons (it stays general rather than explaining each eliminated group), and the reasons it does provide are not grounded in the screenshots. Therefore, the response does not meet the requirement of explaining why each food group is eliminated.
        \item \textbf{Verdict:} NOT SUCCESS
    \end{itemize}

    \item \textcolor{successgreen}{\textbf{Final verdict: NOT SUCCESS}}
\end{itemize}

\section{Prompts} \label{app:prompt}

  \subsection{Agent Action Prompt} \label{app:prompt/agent}

The prompt we use to rollout trajectories is the agent action prompts shown below. Both the official prompt as provided in~\citet{qwen3vl-repo} and the memory prompt that we use across~\S\ref{sec:train} are included below.

\begin{tcblisting}{
  title={Prompt: Agent Action Generation},
  listing only, breakable,
  left=2mm, right=2mm, top=1mm, bottom=1mm,
  listing options={style=promptroman},
}
================================================================================
  MESSAGE 1: SYSTEM
  ================================================================================
  You are a helpful assistant.

  # Tools

  You may call one or more functions to assist with the user query.

  You are provided with function signatures within <tools></tools> XML tags:
  <tools>
  {
    "name": "computer_use",
    "description": "Use a mouse and keyboard to interact with a computer. The screen's resolution is 1000x1000.
  * You do not have access to download files or play videos.
  * Focus on web browsing and navigation tasks only.",
    "parameters": {
      "type": "object",
      "properties": {
        "action": {
          "type": "string",
          "description": "The action to perform:
  * `left_click`: Click the left mouse button at the specified coordinates.
  * `type`: Type text at the specified coordinates. The system will automatically click at the coordinates, type the text, and press Enter.
  * `scroll`: Scroll the page in the specified direction (up or down).
  * `wait`: Wait for the specified number of seconds for changes to occur.
  * `go_back`: Go back to the previous page in browser history.
  * `navigate`: Navigate directly to a specific website URL. The URL must start with https://. CRITICAL: If you see reCAPTCHA or any CAPTCHA challenge on the screen, DO NOT attempt to solve it. Instead, immediately use the navigate action to go to a different relevant website to complete your task.
  * `answer`: Provide the final answer to complete the task.",
          "enum": ["left_click", "type", "scroll", "wait", "go_back", "navigate", "answer"]
        },
        "coordinate": {
          "type": "array",
          "description": "[x, y] coordinates (0-1000 range). Required for left_click and type actions. For type action, specify WHERE to type (e.g., coordinates of input field).",
          "items": {
            "type": "integer",
            "minimum": 0,
            "maximum": 1000
          },
          "minItems": 2,
          "maxItems": 2
        },
        "text": {
          "type": "string",
          "description": "Text to type or answer. Required for type and answer actions. Note: For type action, the system will automatically click at the coordinates, type the text, and press Enter - no need to click separately before typing."
        },
        "direction": {
          "type": "string",
          "enum": ["up", "down"],
          "description": "Scroll direction. Required for scroll action."
        },
        "time": {
          "type": "number",
          "description": "Seconds to wait. Required for wait action."
        },
        "url": {
          "type": "string",
          "description": "URL to navigate to. Required for navigate action. Must start with https://."
        }
      },
      "required": ["action"]
    }
  }
  </tools>

  For each function call, return a JSON object with function name and arguments within <tool_call></tool_call> XML tags:
  <tool_call>
  {"name": <function-name>, "arguments": <args-json-object>}
  </tool_call>

  # Response format

  {{if w/ Memory Prompt}}
    Response format for every step:
    1) Memory: facts you would like to memorize for future actions in json format. Include the current step.
    2) Progress: Decompose the task into subtasks and what has been finished so far with json format. Include progress of the current step.
    3) Intention: clearly state which subtask you're working on at this step with the json key.
    4) Action: a short sentence describing what to do in the UI to accomplish the next subtask.
    5) A single <tool_call>...</tool_call> block containing only the JSON: {"name": <function-name>, "arguments": <args-json-object>}.
    
    Rules:
    - Output exactly in the order: Memory, Progress, Intention, Action, <tool_call>.
    - You MUST use json format for the Memory and Progress parts.
    - Example Task: "Search and compare the prices and locations of product 1 and product 2 on Amazon."
      - Example of Memory json format: {"Price of product 1": "10.00", "Location of product 1": "10.00", "Price of produce 2": "12.00"}.
      - Example of Progress json format: {"Go to Amazon.com": "finished", "Search for price of product 1": "finished", "Search for location of product 1": "finished", "Search for price of product 2": "finished", "Search for location of product 2": "not finished", "Compare product 1 and product 2": "not finished"}.
      - Example of Intention json key format: "Search for location of product 2".
    - You CAN NOT modify previous Memory. Only append to it.
    - You CAN modify Progress from previous conversation to further decompose the task and guide your next action.
      - For example, if the previous assistant message specifies Progress: {"Go to Amazon.com": "finished", "Search for product 1": "finished", "Search for product 2": "not finished", "Compare product 1 and 2": "not finished"},
      - You should further decompose "Search for product 1" and "Search for product 2" into "Search for price of product 1" and "Search for location of product 1", and "Search for price of product 2" and "Search for location of product 2".
    - Do not output anything else outside those five parts."""

  {{/if w/ Memory Prompt}}

  {{if w/o Memory Prompt}}
 
  Response format for every step:
  1) Action: a short sentence describing what to do in the UI.
  2) A single <tool_call>...</tool_call> block containing only the JSON: {"name": <function-name>, "arguments": <args-json-object>}.

  Rules:
  - Output exactly in the order: Action, <tool_call>.
  - Action describes the high-level intention of the tool call within a single sentence.
  - Do not output anything else outside those two parts.
  
  {{/if w/o Memory Prompt}}

================================================================================
  MESSAGE 2: USER
  ================================================================================
  [IMAGE: Screenshot of current webpage]
  Please generate the next action according to the UI screenshot and task.

  Task: Find the details of the public transport journey from Glasgow Queen Street to Edinburgh Waverley, departing on May 22, 2025, at 09:00 am.

  Initial website: https://www.nationalrail.co.uk

  Generate the next action to complete the task.
  ================================================================================
  MESSAGE 3: ASSISTANT {{Example response, note that when using the w/ memory prompt, the previous assistant message will contain memory response}}
  ================================================================================
  Action: Click on the search input field to enter the departure station.

  <tool_call>
  {"name": "computer_use", "arguments": {"action": "left_click", "coordinate": [245, 312]}}
  </tool_call>

  ================================================================================
  MESSAGE 4: USER
  ================================================================================
  [IMAGE: Screenshot after clicking]

  ================================================================================
  MESSAGE 5: ASSISTANT {{Example response, note that when using the w/ memory prompt, the previous assistant message will contain memory response}}
  ================================================================================
  Action: Type the departure station name into the search field.

  <tool_call>
  {"name": "computer_use", "arguments": {"action": "type", "coordinate": [245, 312], "text": "Glasgow Queen Street"}}
  </tool_call>
  \end{tcblisting}

\subsection{Evaluator Prompt: Keypoint Detection} \label{app:prompt/eval/keypoint}

\begin{tcblisting}{title={Prompt: Evaluator/Evaluation Criteria Generation},
  listing only, breakable,
  left=2mm, right=2mm, top=1mm, bottom=1mm,
  listing options={style=promptroman}
}
You are an expert evaluator determining whether an image contains relevant information for completing a task.

  **Instructions**:
  - Answer "YES" if the image shows ANY task-related content: actions taken, progress, search results, tool usage, error messages, or blocking screens.
  - Answer "NO" only if completely irrelevant: generic homepage, unrelated webpage, or blank screens with no context.
  - When in doubt, answer "YES".

  **Response format**:
  1. **Reasoning**: [One sentence explanation]
  2. **Decision**: [YES or NO]

  User Prompt:

  **Task**: {task}
  **Key Points for Task Completion**:
  {eval_rubric}

  The snapshot of the web page is shown in the image. Does this image contain relevant information for the task? (Answer YES unless it's completely irrelevant)

  Where:
  - {task} = task.task_name (the task description)
  - {eval_rubric} = List of all criteria from task.evaluator_reference

  Example filled in:
  **Task**: Find the price of iPhone 15 Pro on Amazon
  **Key Points for Task Completion**:
  - The agent should provide the exact price of the iPhone 15 Pro
  - The price must be visible in the screenshot
  - The product must be from Amazon
\end{tcblisting}

\subsection{Evaluator Prompt: Per-criteria Evaluation} \label{app:prompt/eval/evaluation}

The evaluation system uses a two-criterion approach, each checked separately:

\begin{itemize}
    \item Criterion A is checked per-fact (each rubric/fact from task decomposition is evaluated separately).
    \item Criterion B is checked once per task (verifies the agent's final answer against screenshots).
\end{itemize}

  \subsubsection{Criterion A: Fact Verification}

  \begin{tcblisting}{title={Prompt: Evaluator/Criterion A - Fact Verification},
    listing only, breakable,
    left=2mm, right=2mm, top=1mm, bottom=1mm,
    listing options={style=promptroman}
  }
  ================================================================================
  MESSAGE 1: SYSTEM
  ================================================================================
  You are an expert in evaluating the performance of a web navigation agent. The agent is designed to help a human user navigate a website to complete a task. Your goal is to verify whether a SPECIFIC FACT can be confirmed by the provided screenshots.

  As an evaluator, you will be presented with the following components:

  1. Task Instruction: The original task description (provided for CONTEXT ONLY)
  2. Fact Group: A group of related facts decomposed from the task instruction (provided for CONTEXT ONLY)
  3. Fact to Check: A specific fact that you need to verify (THIS IS YOUR PRIMARY FOCUS)
  4. Trajectory: A complete list of observations and actions that were taken by the agent
  5. Result Screenshots: Visual representation of the screen showing the result or intermediate state

  CRITICAL: Your judgment should ONLY focus on whether the FACT TO CHECK can be verified by the screenshots. You are NOT checking the agent's response - only whether the screenshots contain evidence for the fact.

  Guidelines for evaluation:
  -- Your primary responsibility is to assess whether the screenshots contain evidence that verifies the FACT TO CHECK.
  -- The fact to check may involve more than one sub-fact. ALL sub-facts must be verifiable from the screenshots.
  -- If the fact requires specific information (e.g., "concert is in the US or Canada"), the screenshots must show this information.
  -- If the evaluation criteria asks to find a specific item, the screenshots must show that exact item (not a similar one).

  IMPORTANT - Handling "OR" conditions:
  -- When the fact or task contains "OR" (e.g., "best books on cooking OR gardening OR home decor"), satisfying ANY ONE of the alternatives is sufficient for SUCCESS.
  -- Example: If the task is "find best books on cooking OR gardening OR home decor" and the screenshots show best cooking books, this is SUCCESS - the agent does NOT need to find all three.
  -- "OR" indicates alternatives/options, not a requirement to verify all items.

  Response format (you should STRICTLY follow the format):
  1. Analysis: [Describe what evidence you see in the screenshots related to the fact to check]
  2. Verdict: [SUCCESS if the fact is verified by screenshots, NOT SUCCESS otherwise]

  ================================================================================
  MESSAGE 2: USER
  ================================================================================
  ===Your Turn===
  Task Instruction (for context only):
  [task_instruction]

  Fact Group (for context only):
  [fact_group]

  Fact to Check (PRIMARY FOCUS - verify this against the screenshots):
  [fact_to_check]

  Completion history:
  [trajectory]

  Relevant screenshots:
  attached.

  Evaluation: (MUST end with line "2. Verdict: [SUCCESS or NOT SUCCESS]")
  \end{tcblisting}

\subsubsection{Criterion B: Response Verification / Anti-Hallucination}

  \begin{tcblisting}{title={Prompt: Evaluator/Criterion B - Anti-Hallucination Check},
    listing only, breakable,
    shrink break goal=0.5\baselineskip,  % <-- add this
    left=2mm, right=2mm, top=1mm, bottom=1mm,
    listing options={style=promptroman}
  }
  ================================================================================
  MESSAGE 1: SYSTEM
  ================================================================================
  You are an expert in detecting hallucinations in web navigation agent responses. Your goal is to verify whether the agent's FINAL RESPONSE is supported by the provided screenshots.

  As an evaluator, you will be presented with the following components:

  1. Task Instruction: The task the agent was trying to complete
  2. Final Response: The agent's answer/response to the task
  3. Result Screenshots: Visual representation of the screens the agent visited

  CRITICAL: Your job is to check if the agent's response contains information that is NOT shown in the screenshots. Agents frequently hallucinate or make up answers that are not verified by what they actually saw.

  Guidelines for evaluation:
  -- Check whether EVERY claim in the agent's response can be verified by the screenshots.
  -- If the response mentions specific facts (names, numbers, dates, locations, etc.), these MUST be visible in the screenshots.
  -- If the response contains information not shown in ANY screenshot, this is a hallucination - mark as NOT SUCCESS.
  -- If the response is vague or says "I couldn't find the information", check if this matches what the screenshots show.
  -- YOU SHOULD EXPECT THAT THERE IS A HIGH CHANCE THAT THE AGENT WILL MAKE UP AN ANSWER NOT VERIFIED BY THE SCREENSHOTS.

  Response format (you should STRICTLY follow the format):
  1. Claims in response: [List the specific claims/facts in the agent's response]
  2. Screenshot verification: [For each claim, state whether it appears in the screenshots]
  3. Verdict: [SUCCESS if ALL claims are verified by screenshots, NOT SUCCESS if ANY claim is not verified]

  ================================================================================
  MESSAGE 2: USER
  ================================================================================
  ===Your Turn===
  Task Instruction:
  [task_instruction]

  Agent's Final Response:
  [response]

  Screenshots from the agent's session:
  attached.

  Evaluation: (MUST end with line "3. Verdict: [SUCCESS or NOT SUCCESS]")
  \end{tcblisting}

  \subsection{Evaluator Prompt: Blocking Detection} \label{app:prompt/eval/block}

  Additionaly, we introduce a blocking detection step, which is used to identify if failures were due to anti-bot measures vs. agent errors. Those websites that frequently blocks the agent will be excluded from being sampled.

  \begin{tcblisting}{title={Prompt: Evaluator/Blocking Detection},
    listing only, breakable,
    left=2mm, right=2mm, top=1mm, bottom=1mm,
    listing options={style=promptroman}
  }
  ================================================================================
  MESSAGE 1: SYSTEM
  ================================================================================
  You are an expert in analyzing web navigation trajectories to detect if a website has EXPLICITLY and UNRECOVERABLY BLOCKED an automated agent using anti-bot measures.

  CRITICAL DISTINCTION:
  - Blocked = Website explicitly prevents access with UNRECOVERABLE anti-bot measures that the agent CANNOT escape from by navigating
  - NOT Blocked = Agent makes navigation mistakes, clicks wrong elements, or encounters temporary obstacles that could be navigated away from

  EXPLICIT SIGNS OF UNRECOVERABLE BLOCKING (answer YES):
  - CAPTCHA challenges that persist and cannot be bypassed (reCAPTCHA, hCaptcha, image selection puzzles)
  - Cloudflare protection screens that remain stuck ("Checking your browser", "Just a moment", "DDoS protection")
  - Explicit "Access Denied" or "403 Forbidden" error pages that appear consistently
  - "Please verify you are human" prompts that block all access
  - Bot detection pages that explicitly state "automated access detected" and prevent further navigation
  - Security challenge pages that completely prevent any website access

  DO NOT COUNT AS BLOCKING (answer NO):
  - Agent clicked wrong buttons or links (navigation mistakes)
  - Agent cannot find information on a normally accessible page (task difficulty)
  - Page loaded slowly but eventually loaded (performance issue, not blocking)
  - Agent got lost or confused navigating the site (agent error)
  - "Page not found" or 404 errors (broken links, not blocking)
  - Empty search results (legitimate website response, not blocking)
  - Temporary errors that the agent could escape from by going back or navigating elsewhere

  CRUCIAL: Only answer "YES" if you see PERSISTENT, UNRECOVERABLE anti-bot measures that prevent ALL navigation on the website.

  Response format (you MUST strictly follow):
  1. Analysis: [Describe what you see in the screenshots - any EXPLICIT signs of anti-bot blocking measures?]
  2. Blocked: [YES or NO]

  ================================================================================
  MESSAGE 2: USER
  ================================================================================
  ===Your Turn===
  Task:
  [task]

  Trajectory:
  [trajectory]

  Screenshots:
  attached.

  Did the website block the agent? (MUST end with line "2. Blocked: [YES or NO]")
  \end{tcblisting}

 \subsection{Task Set Generation Prompt: Evaluation Criteria Proposal} \label{app:prompt/task/criteria}

  \begin{tcblisting}{title={Prompt: Task/Evaluation Criteria Generation},
    listing only, breakable,
    left=2mm, right=2mm, top=1mm, bottom=1mm,
    listing options={style=promptroman}
  }
  ======== Guidelines ========
  You're a helpful assistant that generates fact-based evaluation rubrics for web navigation tasks.

  Your task is to:
  1. Break down the task into hierarchical fact groups, where each group contains specific facts to verify
  2. Each fact represents ONE verifiable piece of information that can be checked in a trajectory
  3. The overall task difficulty equals the TOTAL NUMBER of all facts across all groups

  Key principles:
  - FIRST analyze the task to identify logical groupings of related information
  - For each group, list ALL specific facts that need to be verified
  - Facts should be AS DETAILED AS POSSIBLE - break down complex requirements into individual checkable facts
  - Each fact has a difficulty of 1 (facts are atomic units)
  - The OVERALL DIFFICULTY is the TOTAL COUNT of all facts across all groups
  - A trajectory is marked correct if and only if ALL facts are verified
  - Do not make up facts not present in the original task

  ======== CRITICAL: Do Not Decompose Mathematical Problems ========

  When a task involves mathematical calculations, physics problems, or computational work, DO NOT break down the calculation steps. Treat the entire computation as a SINGLE fact.

  BAD Example:
  Task: What is the result of 2^3 + 2^2 + 2^1 + 2^0?

  {
    "fact_groups": [{"id": 1, "facts": ["calculate 2^3", "calculate 2^2", ...]}],
    "difficulty": 5
  }
  Problem: Breaking down calculation into individual steps.

  CORRECT:
  {
    "fact_groups": [{"id": 1, "facts": ["result of the equation 2^3 + 2^2 + 2^1 + 2^0"]}],
    "difficulty": 1
  }

  ======== CRITICAL: Do Not Decompose "How-To" Instructions ========

  When a task asks "How to" do something, the task is asking for INSTRUCTIONS from the website, not to verify each step was performed.

  BAD Example:
  Task: How can I set up an Apple Watch for the first time?

  {
    "fact_groups": [
      {"id": 1, "facts": ["ensure Apple Watch is charged", "check compatibility"]},
      {"id": 2, "facts": ["turn on Apple Watch", "bring close to iPhone", ...]}
    ],
    "difficulty": 11
  }
  Problem: Breaking down procedure into action steps.

  CORRECT:
  {
    "fact_groups": [{"id": 1, "facts": ["instructions for setting up Apple Watch for the first time"]}],
    "difficulty": 1
  }

  ======== CRITICAL: AND vs OR Logic ========

  ALL facts must be verified - facts use AND logic by default.
  If the task contains OR statements (e.g., "do X or Y"), represent as a SINGLE fact with OR capitalized.

  BAD Example:
  Task: What should a user do if they were trying to contact the original owners or access the previous content.

  {
    "fact_groups": [
      {"id": 1, "facts": ["method to contact the original owners"]},
      {"id": 2, "facts": ["method to access the previous website content"]}
    ]
  }
  Problem: Treating OR as AND (creating separate groups).

  CORRECT:
  {
    "fact_groups": [
      {"id": 1, "facts": ["method to contact the original owners OR access the previous website content"]}
    ],
    "difficulty": 1
  }

  ======== CRITICAL: Do Not Enumerate Unspecified Items ========

  When a task asks to "identify" or "find" items WITHOUT specifying what they are, DO NOT make up specific facts. Similarly, do not enumerate "details" or "information" into specific types.

  BAD Example:
  Task: What are the contact details for Hamble Harbour Office?

  {
    "fact_groups": [{"id": 1, "facts": ["telephone number", "email address", "physical address"]}],
    "difficulty": 3
  }
  Problem: Enumerating specific types when task just says "contact details".

  CORRECT:
  {
    "fact_groups": [{"id": 1, "facts": ["contact details for Hamble Harbour Office"]}],
    "difficulty": 1
  }

  ======== CRITICAL: "Such as" Are Illustrative, Not Facts ========

  Phrases like "such as", "for example", "e.g." provide examples to clarify, NOT additional facts.

  BAD Example:
  Task: Find a vegetarian lunch dish, such as chickpea salad or lentil soup, with 5-star rating.

  {
    "fact_groups": [{"id": 1, "facts": ["recipe is vegetarian", "example dishes include chickpea salad or lentil soup"]}]
  }
  Problem: Creating fact for examples.

  CORRECT:
  {
    "fact_groups": [{"id": 1, "facts": ["recipe is for a vegetarian dish", "dish is suitable for lunch", "recipe has a 5-star rating"]}],
    "difficulty": 3
  }

  ======== Example ========

  Task: Find the full list of exhibitions at The Ringling, including their titles, dates, and descriptions.

  Expected response:
  {
    "fact_groups": [
      {
        "id": 1,
        "description": "find complete list of exhibitions at The Ringling",
        "facts": [
          "full list of exhibitions",
          "current exhibitions included",
          "upcoming exhibitions included",
          "exhibitions are at The Ringling"
        ]
      },
      {
        "id": 2,
        "description": "get details for each exhibition",
        "facts": [
          "title for each exhibition",
          "specific dates for each exhibition (if available)",
          "brief description for each exhibition"
        ]
      }
    ],
    "difficulty": 7
  }

  ======== Output Format ========
  Respond with valid JSON:
  - "fact_groups": Array of objects with "id", "description", "facts" (array of strings)
  - "difficulty": Integer = TOTAL COUNT of all facts across all groups

  ======== Your Turn ========
  Task: {task}
  >>>>>>>>>>>
  \end{tcblisting}

  \subsection{Task Set Generation Prompt: Task Proposal from Fact Groups} \label{app:prompt/task/task}

  \begin{tcblisting}{title={Prompt: Task/Task Generation from Selected Fact Groups},
    listing only, breakable,
    shrink break goal=0.5\baselineskip,  % <-- add this
    left=2mm, right=2mm, top=1mm, bottom=1mm,
    listing options={style=promptroman}
  }
  ======== Task Generation ========

  You are a helpful assistant that generates simplified task descriptions based on selected fact groups from the original task.

  Original Task: {original_task}

  Selected Fact Groups:
  {selected_groups}

  Your task is to:
  1. Create a new task description that incorporates ONLY the requirements explicitly listed in the selected fact groups above
  2. The new task should read naturally and be self-contained
  3. Do NOT include requirements from fact groups that were not selected
  4. Maintain the original context and domain of the task
  5. **CRITICAL**: Do NOT add any information that is not explicitly present in the selected fact groups above, even if that information appears in the original task

  Important Guidelines:
  - If a fact group mentions "concert" but does NOT mention "upcoming" or "in the US or Canada", do NOT include those constraints in the generated task
  - ONLY use constraints and requirements that are explicitly listed as facts in the selected groups
  - Do not infer or add information from the original task that is not in the selected fact groups

  Output Format:
  Respond with ONLY the new task description as plain text (no JSON, no extra formatting).

  Example:
  Original Task: "Find the full list of current and upcoming exhibitions at The Ringling, including their titles, specific dates (if available), and a brief description for each."

  Selected Fact Groups:
  - Group 1 (find complete list of exhibitions at The Ringling): full list of exhibitions, current exhibitions included, upcoming exhibitions included, exhibitions are at The Ringling

  New Task: "Find the full list of current and upcoming exhibitions at The Ringling."

  Your Turn:
  >>>>>>>>>>>
  \end{tcblisting}

\section{Hyper-parameters}
\label{app:hparams}

Our system consists of four main components: environment management, data collection (rollout), model training, and the agent model. The hyperparameters for each component are carefully
tuned to balance performance, scalability, and sample efficiency. The final hyperparameters can be found in Table~\ref{tab:hyperparameters}.

\begin{table}[H]
  \caption{\textbf{Hyperparameters for All Experiments}. Bolded option suggests an experimented optimality.}
  \centering
  \begin{tabular}{lcc}
  \toprule
  Component & Hyperparameter & Value \\
  \midrule
  \multirow{5}{*}{\textbf{Environment (2 workers)}}
  & Simulation CPU Server size & 80 \\
  & HTTP pool size (screenshot/execute) & 256 \\
  & HTTP pool size (metadata/navigate) & 256 \\
  & HTTP pool size (allocate/release) & 4 \\
  & Max vLLM sessions & 80 \\
  \midrule
  \multirow{7}{*}{\textbf{Rollout}}
  & Train tasks rollout size & 512, \textbf{1024} \\
  & Task Sampling & \textbf{uniform}/ratio \\
  & Max steps (easy,medium,hard) & (15,30,45), \textbf{(10,20,30)} \\
  & Interaction mode & \textbf{coordinates-only}, set-of-marks \\
  & Temperature & 1.0 \\
  & Top-p & 0.99 \\
  & Top-k & 2 \\
  \midrule
  \multirow{13}{*}{\textbf{Training}}
  & Learning rate & 1e-5, 5e-6, \textbf{1e-6} \\
  & Effective batch size & 3*4*8*2 \\
  & Times each sample is trained & $\approx 2$ \\
  & Max gradient norm & 1.0 \\
  & Weight decay & 0, \textbf{0.01} \\
  & Warmup steps & \textbf{30}, 100 \\
  & LR scheduler type & \textbf{constant\_with\_warmup}, cosine \\
  & Samples to train per iteration & 1200, \textbf{1800} \\
  & Recency bias power & 1, \textbf{2} \\
  & Replay buffer capacity & \textbf{last 4}, last 8 \\
  & Cutoff length & 4096, 8192, \textbf{16384} \\
  & Gradient checkpointing & False \\
  & bf16 precision & True \\
  \midrule
  \multirow{3}{*}{\textbf{Model}}
  & Model type & Qwen3-VL-8B \\
  & Variant & \textbf{Instruct, Thinking} \\
  & Max new tokens & 1024, 2048, \textbf{3072} \\
  \bottomrule
  \end{tabular}
  \label{tab:hyperparameters}
  \end{table}

\end{document}